\documentclass{article}

\PassOptionsToPackage{numbers,sort&compress,square}{natbib}

\usepackage[preprint]{neurips_data_2024}

\usepackage[utf8]{inputenc} 
\usepackage[T1]{fontenc}    
\usepackage{hyperref}       
\usepackage{url}            
\usepackage{booktabs}       
\usepackage{amsfonts}       
\usepackage{nicefrac}       
\usepackage{microtype}      
\usepackage{xcolor}         

\usepackage{wrapfig}
\usepackage{caption}
\usepackage{multirow}
\usepackage{graphicx}
\usepackage{subfigure}
\usepackage{tabularx}
\usepackage{amsmath}

\usepackage[page]{appendix}
\usepackage{etoolbox}
\usepackage{hyperref}
\usepackage{lipsum}
\renewcommand{\appendixtocname}{Contents}

\makeatletter
\let\oldappendix\appendices

\g@addto@macro\tableofcontents{%
  \let\tf@toc@orig\tf@toc
}
\renewcommand{\appendices}{%
  \clearpage
  \renewcommand{\thesection}{\Roman{section}}
  \let\tf@toc\tf@app
  \addtocontents{app}{\protect\setcounter{tocdepth}{2}}
  \immediate\write\@auxout{%
    \string\let\string\tf@toc\string\tf@app
  }
  \oldappendix
}%

\g@addto@macro\endappendices{%
  \let\tf@toc\tf@toc@orig
  \immediate\write\@auxout{%
    \string\let\string\tf@toc\string\tf@toc@orig
  }%
}  

\newcommand{\listofappendices}{%
  \begingroup
  \renewcommand{\contentsname}{\appendixtocname}
  \let\@oldstarttoc\@starttoc
  \def\@starttoc##1{\@oldstarttoc{app}}
  \tableofcontents
  \endgroup
}

\makeatother

\title{Younger: The First Dataset for Artificial Intelligence-Generated Neural Network Architecture}

\author{%
  \textbf{Zhengxin Yang}$^{1,2}$ \quad \textbf{Wanling Gao}$^{1,2}$ \quad \textbf{Luzhou Peng}$^{1,2}$ \\
  \textbf{Yunyou Huang}$^{3}$ \quad \textbf{Fei Tang}$^{4,5}$ \quad \textbf{Jianfeng Zhan}$^{1,2}$\thanks{Corresponding author} \\
  Research Center of Distributed Systems,\\
  Institute of Computing Technology, Chinese Academy of Sciences $^{1}$\\
  University of Chinese Academy of Sciences $^{2}$\\
  Key Lab of Education Blockchain and Intelligent Technology,\\
  Ministry of Education, Guangxi Normal University $^{3}$\\
  Inspur Group Co., Ltd. $^{4}$ \\
  Jinan Inspur Data Technology Co., Ltd. $^{5}$\\
  \texttt{\{yangzhengxin17z, gaowanling, zhanjianfeng\}@ict.ac.cn}$^{1,2}$ \\
  \texttt{huangyunyou@gxnu.edu.cn}$^{3}$ \texttt{tangfei01@inspur.com}$^{4,5}$ \\
}

\begin{document}

\maketitle

\begin{abstract}
Designing and optimizing neural network architectures typically requires extensive expertise, starting with handcrafted designs and then manual or automated refinement. This dependency presents a significant barrier to rapid innovation. Recognizing the complexity of automatically generating neural network architecture from scratch, we introduce Younger, a pioneering dataset to advance this ambitious goal. Derived from over 174K real-world models across more than 30 tasks from various public model hubs, Younger includes 7,629 unique architectures, and each is represented as a directed acyclic graph with detailed operator-level information. The dataset facilitates two primary design paradigms: global, for creating complete architectures from scratch, and local, for detailed architecture component refinement. By establishing these capabilities, Younger contributes to a new frontier, Artificial Intelligence-Generated Neural Network Architecture (AIGNNA). Our experiments explore the potential and effectiveness of Younger for automated architecture generation and, as a secondary benefit, demonstrate that Younger can serve as a benchmark dataset, advancing the development of graph neural networks. We release the dataset\footnote{Younger Dataset: \url{https://datasets.yangs.cloud/younger}} and code\footnote{Source Code: \url{https://github.com/Jason-Young-AI/Younger}} publicly to lower the entry barriers and encourage further research in this challenging area.
\end{abstract}


\section{Introduction}\label{introduction}
The proliferation of large language models like ChatGPT~\cite{openaiGPT4TechnicalReport2023} has decisively demonstrated the critical importance of large-scale data collection and innovative neural network architecture design in advancing Artificial Intelligence (AI)~\cite{schuhmannLAION5Bopenlargescale2022,tangAGIBenchMultigranularityMultimodal2024}. Mainstream models have shifted from architectures predominantly based on RNNs or CNNs to those employing the Transformer~\cite{vaswaniAttentionAllYou2017}, consistently delivering superior performance across various tasks~\cite{longVisionandLanguagePretrainedModels2022,zhouComprehensiveSurveyPretrained2023}. This shift illustrates the significant impact of architectural innovations. It compels us to investigate Artificial Intelligence-Generated Neural Network Architecture (AIGNNA), a concept we introduce for the autonomous generation of neural architectures from scratch, as shown in Figure~\ref{fig:all_show}.

Designing and optimizing neural network architectures requires profound domain expertise, meticulous manual construction, and continuous refinement. Such demands underscore the challenges in achieving rapid and automated generation of efficient neural network architectures. While Neural Architecture Search (NAS) has achieved significant success in automating the optimization and search of neural network architectures~\cite{tanEfficientNetRethinkingModel2019a,tanMnasNetPlatformAwareNeural2019,kleinTabularBenchmarksJoint2019}, it still heavily relies on human expertise for designing the initial overall network topology or macro-architecture~\cite{anDiffusionNAGPredictorguidedNeural2023,youDesignSpaceGraph2021,yingNASBench101ReproducibleNeural2019}. This reliance restricts the diversity and innovation of neural network architectures.

To overcome the reliance on human expertise and foster the automated generation of diverse and innovative neural architectures, we introduce Younger, a pioneering dataset that encapsulates extensive prior knowledge about neural network architecture. Derived from approximately 174K real-world models across more than 30 tasks from various public model repositories, as listed in Table~\ref{tab:model_sources}, Younger includes 7,629 distinct neural network architectures, each represented as a directed acyclic graph (DAG) with detailed operator-level information. Specifically, all models are converted into the Open Neural Network Exchange (ONNX) format~\cite{bai2019onnx}, an open format designed to represent neural network models, ensuring the standardization of architecture and framework independence. To ensure model uniqueness and address privacy and safety concerns, we developed a tool that converts ONNX models into DAGs while preserving architectural integrity. Parameter values are excluded to ensure privacy and safety. In these DAGs, nodes represent ONNX operators, recording detailed configurations and hyperparameters, while edges represent the data flow between operators.

Activated by the Younger, we introduce two paradigms for AIGNNA, characterized by escalating complexity, as shown in Figure~\ref{fig:all_show}: 1) Local: This paradigm focuses on fine-tuning and optimizing components of pre-existing or pre-generated neural network architectures. Within the paradigm, the challenge intensifies from determining data flows between operators to select the most suitable operator type for the given architecture. Due to the large number of operator types, the selection and integration of operators present a higher level of difficulty. 2) Global: This more demanding paradigm entails generating entire neural network architectures from scratch with no restrictions, epitomizing an end-to-end automation challenge that requires building a complete and functional architecture starting from zero.

We conducted experiments to validate the Younger's potential and effectiveness for Artificial Intelligence-Generated Neural Network Architecture. Initially, we performed statistical analyses at three different levels of granularity: operator level, component level, and architecture level, demonstrating that the dataset contains sufficiently rich prior knowledge. Subsequently, we explored the practical application of the local and global paradigms, proving the dataset's feasibility in real-world scenarios. Furthermore, our experiments revealed that due to the representation of architecture as DAG, Younger's unique graph structural characteristics make it a promising \textbf{benchmark dataset in advancing research in Graph Neural Networks} (GNN).

As the name of our dataset, `Younger,' implies, we have made the dataset publicly available online to keep it perpetually young and to lower the barriers to entry for research, thus fostering development in the field. To promote the democratization of research, we also provide a platform that allows researchers worldwide to upload their models. Our platform can automatically convert the uploaded models into DAG format and integrate them into the next release of Younger.


\section{Related Work}\label{related_work}
\subsection{Artificial Intelligence-Generated Neural Network Architecture}
The design and optimization of neural network architectures have historically been labor-intensive tasks, relying heavily on the intuition and expertise of human researchers. This process has evolved from manual designs, exemplified by early architectures such as AlexNet\cite{krizhevskyImageNetClassificationDeep2012}, ResNet \cite{heDeepResidualLearning2016}, LSTM \cite{neco}, and Transformer \cite{vaswaniAttentionAllYou2017}, to more automated methods employing neural architecture search (NAS) like NASNet~\cite{zoph2018learning} and DARTS~\cite{liuDARTSDifferentiableArchitecture2018}.

Early architecture designs, while innovative, were constrained by their reliance on expert knowledge and extensive experience. Despite the effectiveness of these manually designed architectures, they needed more agility and speed in an era characterized by rapid data growth and increasingly diverse application domains. This temporal inefficiency made it challenging to quickly develop tailored neural network solutions that could adapt to and capitalize on emerging opportunities across varied fields.

The advent of NAS marked a significant shift towards automation, aimed at enhancing the efficiency and adaptability of the design process. NAS frameworks like DARTS \cite{liuDARTSDifferentiableArchitecture2018} and benchmarks such as NAS-Bench-101 \cite{yingNASBench101ReproducibleNeural2019} and NATS-Bench \cite{dongNATSBenchBenchmarkingNAS2021} introduced methodologies that automate the exploration of predefined search spaces (macro-architectures) and the selection of optimal cells (building blocks or micro-architectures)~\cite{zoph2018learning,aaai.v33i01.33014780,978-3-030-01246-5_2,liuDARTSDifferentiableArchitecture2018,pmlr-v80-pham18a,tanEfficientNetRethinkingModel2019a,kleinTabularBenchmarksJoint2019}. However, the potential for innovation in NAS is constrained by the limitations of predefined macro-architectures and a restricted selection of operators, which stifles the exploration of new architectural designs.

\begin{table}[!htpb]
    \centering
    \begin{tabular}{c||ccc|c|c}
    \hline
    \multirow{2}{*}{\textbf{Dataset}} &
      \multicolumn{3}{c|}{\textbf{\#architectures}} &
      \multirow{2}{*}{\textbf{\#op-types}} &
      \multirow{2}{*}{\textbf{\#tasks}} \\ \cline{2-4}
     &
      \multicolumn{1}{c|}{\textbf{unrestricted-}} &
      \multicolumn{1}{c|}{\textbf{macro-}} &
      \textbf{micro-} &
       &
       \\ \hline \hline
    NAS-Bench-101~\cite{yingNASBench101ReproducibleNeural2019} & \multicolumn{1}{c|}{-}    & \multicolumn{1}{c|}{1} & 423K & 3 (Desc.)  & 1 (Image)      \\ \hline
    NAS-Bench-201~\cite{dongNASBench201ExtendingScope2019} & \multicolumn{1}{c|}{-}    & \multicolumn{1}{c|}{1} & 15K  & 5 (Desc.)  & 1 (Image)      \\ \hline
    NAS-Bench-NLP~\cite{klyuchnikovNASBenchNLPNeuralArchitecture2022} & \multicolumn{1}{c|}{-}    & \multicolumn{1}{c|}{1} & 14K  & 6 (Desc.)  & 1 (Text)       \\ \hline
    NAS-Bench-ASR~\cite{mehrotraNASBenchASRReproducibleNeural2020} & \multicolumn{1}{c|}{-}    & \multicolumn{1}{c|}{1} & 8K   & 6 (Desc.)  & 1 (Audio)      \\ \hline
    DeepNets-1M~\cite{knyazevParameterPredictionUnseen2021}   & \multicolumn{1}{c|}{-}    & \multicolumn{1}{c|}{1} & 1M   & 15 (Desc.) & 1 (Image)      \\ \hline
    \textbf{Younger}       & \multicolumn{1}{c|}{7.4K} & \multicolumn{1}{c|}{-} & -    & 193 (ONNX) & 31 (Unlimited) \\ \hline
    \end{tabular}
    \caption{\textbf{The difference between Younger and NAS-related datasets}. Younger does not restrict macro- or micro-architectures, it contains rich real-world architectures.}
    \label{tab:dataset_nas}
\end{table}%
The Younger dataset and the associated AIGNNA methodology propose a revolutionary departure from these constraints. By eliminating the need for predefined macro-architectures, Younger allows for a more explorative approach to architecture design, supporting various operator types and data flow configurations, as seen in Table~\ref{tab:dataset_nas}. This flexibility facilitates the generation of innovative, customized architectures better suited to specific applications and more adaptable to emerging challenges in neural network design.

\subsection{Benchmarking Graph Neural Network}
Graph neural networks (GNNs) have gained prominence due to their effectiveness in processing data represented in graph formats, applicable across various domains such as social network analysis, recommendation systems, and molecular chemistry. Standard benchmark datasets for GNN research include citation networks like Cora, CiteSeer, and PubMed \cite{pmlr-v48-yanga16}, as well as molecular datasets such as QM9 \cite{C7SC02664A} and ZINC \cite{doi:10.1021/acscentsci.7b00572}. These datasets primarily represent graphs with relatively simple and static structures.

\begin{wraptable}{r}{0.7\textwidth}
  \centering
  \begin{tabular}{c||c|c|c|c}
  \hline
  \textbf{Dataset} &
    \textbf{\#graphs} &
    \textbf{\#nodes} &
    \textbf{\#edges} &
    \textbf{\#node-types} \\ \hline\hline
  Cora~\cite{pmlr-v48-yanga16} &
    1 &
    2,708 &
    10,556 &
    - \\ \hline
  CiteSeer~\cite{pmlr-v48-yanga16} &
    1 &
    3,327 &
    9,104 &
    - \\ \hline
  PubMed~\cite{pmlr-v48-yanga16} &
    1 &
    19,717 &
    88,648 &
    - \\ \hline
  ZINC~\cite{doi:10.1021/acscentsci.7b00572} &
    49,456 &
    $\sim$23.2 &
    $\sim$49.8 &
    10 \\ \hline
  QM9~\cite{C7SC02664A} &
    130,831 &
    $\sim$18.0 &
    $\sim$37.3 &
    5 \\ \hline
  \textbf{Younger} &
    7,629 &
    $\sim$1,658 &
    $\sim$2,113 &
    184 \\ \hline
  \end{tabular}
  \caption{\textbf{The difference between Younger and GNN-related datasets}. Younger has the highest number of node types and is more balanced than other datasets regarding the number of nodes, edges, and graphs.}
  \label{tab:dataset_gnn}
\end{wraptable}%
Due to the inherent nature of the directed acyclic graph of the data items in Younger, it can naturally be used as a benchmark data set for graph neural network-related research. As shown in Table~\ref{tab:dataset_gnn}, Younger introduces a new dimension to GNN benchmarking by providing a dataset of diverse, complex directed acyclic graphs (DAGs) ranging dramatically in scale and complexity. This variability presents new challenges and opportunities for GNN research, particularly in testing the scalability, robustness, and generalizability of GNN algorithms. Each graph within Younger can be studied as homogeneous or heterogeneous, depending on the specific research focus or application, providing a versatile platform for advancing GNN methodologies. Homogeneous or heterogeneous considerations are discussed in Section~\ref{experiments}.

Developing the Younger dataset and the AIGNNA methodology significantly advances neural network architecture design and GNN benchmarking. By providing paradigms that support extensive customization and innovation, Younger facilitates a deeper exploration of neural architectures and offers a robust benchmarking tool for GNN research. This approach challenges existing approaches and paves the way for future neural network design developments.


\section{Dataset Construction}\label{construction}
Collecting real-world neural network architectures is complicated and requires in-depth knowledge of deep learning frameworks, particularly ONNX~\cite{bai2019onnx}, as well as significant human effort, computing power, and network resources. This can be prohibitive for many researchers. To provide broad support for AIGNNA and lower the entry barriers for researchers, we provide a series of neural network architecture collection and processing tools to avoid the labor costs required for data collection. Using these tools, we constructed the initial version of Younger, with resource consumption detailed in the Appendix.
Our dataset construction process mainly includes four parts: retrieving neural network models, converting models to ONNX, extracting DAGs from ONNX, and filtering out unique DAGs. Figure~\ref{fig:dataset_construction} shows the overall processing pipeline. The specific process is as follows:

\begin{figure}[!htbp]
\centering
\includegraphics[width=0.93\textwidth]{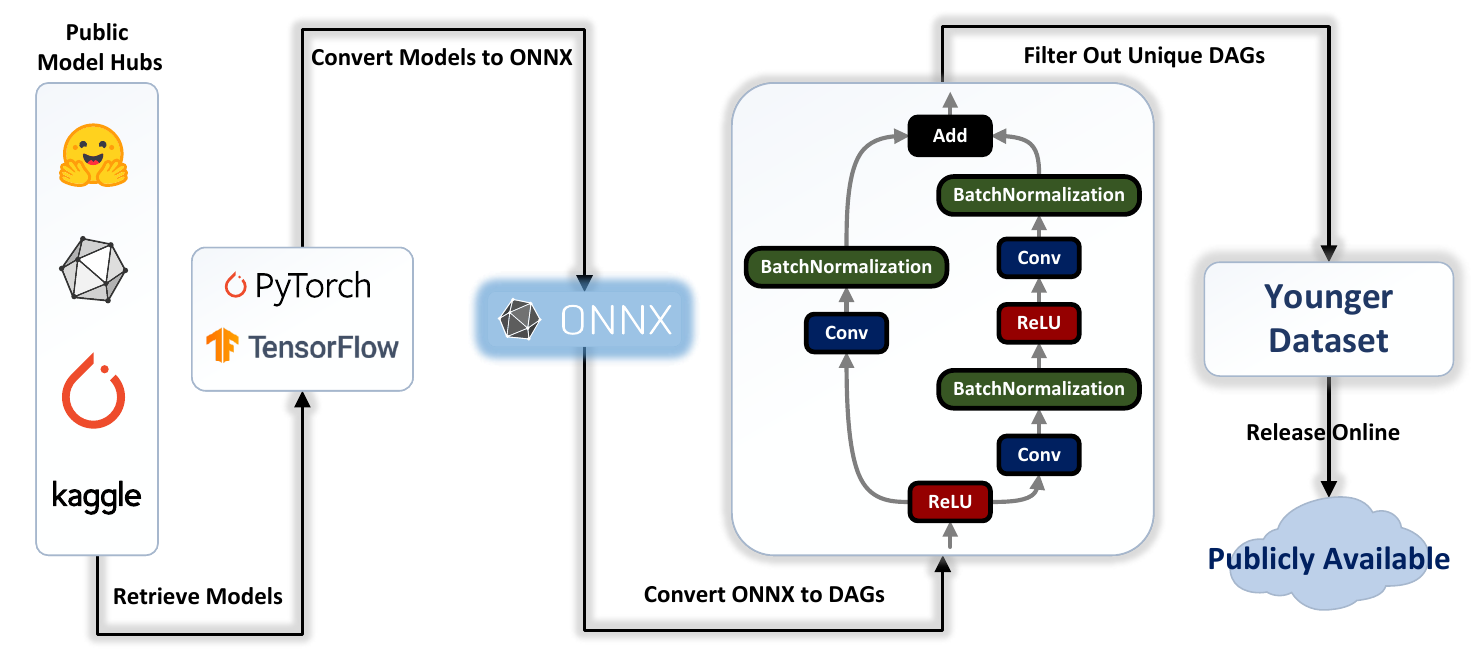}
\caption{\textbf{Overview of the construction pipeline}. Models are retrieved from Hugging Face Hub, ONNX Model Zoo, PyTorch Hub, and Kaggle Models. Then, all retrieved models are converted to ONNX and extracted to be DAGs.}
\label{fig:dataset_construction}
\end{figure}
\vspace{-10pt}

\textbf{Retrieving Neural Network Models:}
We constructed Younger using four well-known neural network models from four publicly available neural network model repositories. Specifical, four well-known open-source neural network model repositories are chosen to obtain a sufficiently diverse range of neural network models: Kaggle Models\footnote{Kaggle Models: \url{https://www.kaggle.com/models}}, PyTorch Hub\footnote{PyTorch Hub: \url{https://pytorch.org/hub/}}, ONNX Model Zoo\footnote{ONNX Model Zoo: \url{https://onnx.ai/models/}}, and Hugging Face Hub\footnote{Hugging Face Hub: \url{https://huggingface.co/models}}. These repositories cover over 30 types of deep learning tasks and include models based on various deep learning frameworks such as PyTorch~\cite{paszkePyTorchImperativeStyle2019} and TensorFlow~\cite{abadiTensorFlowsystemlargescale2016}, see Appendix for details. It ensures the diversity of sources for deep learning models to a certain extent. Given the rapid growth rate of the Hugging Face Hub, we have provided a series of automated model acquisition tools to support Younger's rapid version iteration and maintain timeliness. In addition, although the update frequency of the other three model libraries is slow, we have also developed corresponding tools to achieve automation. Table~\ref{tab:model_sources} shows detailed information about the model hubs.

\textbf{Converting Models to ONNX:}
Different deep learning frameworks define different operators. Without a unified representation, issues such as increased dataset usage costs and reduced efficiency in architecture design may arise. Therefore, we adopt Open Neural Network Exchange (ONNX) as the unified representation format for the model. ONNX defines machine learning models by providing a standard set of operators, allowing different deep learning frameworks (such as PyTorch, TensorFlow, etc.) to use the same operators for model exchange and deployment.
Another benefit of using ONNX as a unified representation is that it significantly reduces the number of operator types, greatly lowering the cost of representing neural network architectures. For instance, while PyTorch has defined approximately 2000+ different operators, with ONNX, only about 200 standard operators are required.
We use several open-source tools for conversion, such as optimum\footnote{Optimum: \url{https://github.com/huggingface/optimum}} and tf2onnx\footnote{tf2onnx: \url{https://github.com/onnx/tensorflow-onnx}}; see Appendix for details.

\textbf{Extracting DAGs From ONNX:}
To protect parameter information and avoid security and privacy issues, all parameter information must be removed from the ONNX model. In addition, compared to DAGs, ONNX models defined in the Protocol Buffers\footnote{Protocol Buffers \url{https://protobuf.dev/}} format are less suitable for direct reading and processing by various data analysis and processing tools like graph studying tools(e.g., NetworkX~\cite{Hagberg2008ExploringNS}) and deep learning frameworks (e.g., PyTorch Geometric~\cite{feyFastGraphRepresentation2019}). Consequently, we have developed a tool for transforming the ONNX models into directed acyclic graphs. This tool allows model owners to provide their architecture design without leaking parameter information, helping enrich the construction and development of open-source Younger.

Specifically, we store each operator in a neural network architecture in the nodes of the DAG and record detailed information such as the operator type and operator attribute definitions. We use directed edges to represent the data flows within the neural network architecture and meticulously document each node's data inflow and outflow order. We utilize the open-source graph library NetworkX to represent the DAGs.

\textbf{Filtering the Dataset:}
There are many isomorphic neural network architectures in public model hubs, so it is necessary to filter the isomorphic architectures to ensure the uniqueness of each architecture in the dataset. We use the Weisfeiler Lehman (WL)~\cite{shervashidzeWeisfeilerLehmanGraphKernels2011} graph hash algorithm to compute the extracted DAGs and identify heterogeneous architectures. The WL algorithm has the same calculation result for isomorphic graphs and firmly guarantees that heterogeneous graphs have different hashes. We use operator type and operator attributes represented in nodes as the iteration objects of the WL hash algorithm, ensuring the heterogeneity of all architectures in the dataset in terms of hyperparameters and operator types. After filtering, we obtained 7629 heterogeneous neural network architectures from 174K real-world models.

\begin{table}[!htpb]
\centering
\begin{tabular}{c||c|c|c|c|c}
\hline
\textbf{Public Model Hubs} & \textbf{Retrievable} & \textbf{Convertable} & \textbf{Retrieved} & \textbf{Converted} & \textbf{Filtered} \\ \hline\hline
Hugging Face Hub & 691K   & 325K & 143.5K & 96K  & \multirow{4}{*}{-} \\ \cline{1-5}
ONNX Model Zoo          & 12K    & 12K  & 12K    & 74K  &                    \\ \cline{1-5}
PyTorch Hub             & N/A    & 121  & 121    & 121  &                    \\ \cline{1-5}
Kaggle Models           & 5K     & 4K   & 4K     & 4K   &                    \\ \hline
Total                   & 743.5K & 341K & 159.5K & 174K & 7629               \\ \hline
\end{tabular}
\caption{\textbf{Statistical information during the construction process of Younger}. This includes the number of \textbf{retrievable} models from four publicly available model hubs, the number of models that are ONNX \textbf{convertable}, the number of models that have been successfully \textbf{retrieved}, the number of models that have been successfully \textbf{converted} to ONNX, and the number of heterogeneous DAGs that are \textbf{filtered}.}
\label{tab:model_sources}
\end{table}%
\vspace{-13pt}
When the first version of Younger started construction, there were 743.5K publicly available models, of which 341K models could be converted to ONNX format. As of the release of the first version of Younger, we have extracted 174K models for processing. It can be seen that even though the base of deep learning models is large and the growth pace is fast, the actual effective and heterogeneous neural network architecture is less than 1\% of the total number of models. Please take a look at Table 1 for details.
The notably low proportion of heterogeneous neural network architectures reveals that current neural network design methods, both manual and NAS-based, have a limited impact on architectural innovation. This underscores Younger's importance in offering new design approaches for AIGNNA and its critical role in advancing neural network architecture innovation.


\section{Paradigms for AI-Generated Neural Network Architecture (AIGNNA)}
\begin{figure}[htbp]
\centering
\includegraphics[width=0.93\textwidth]{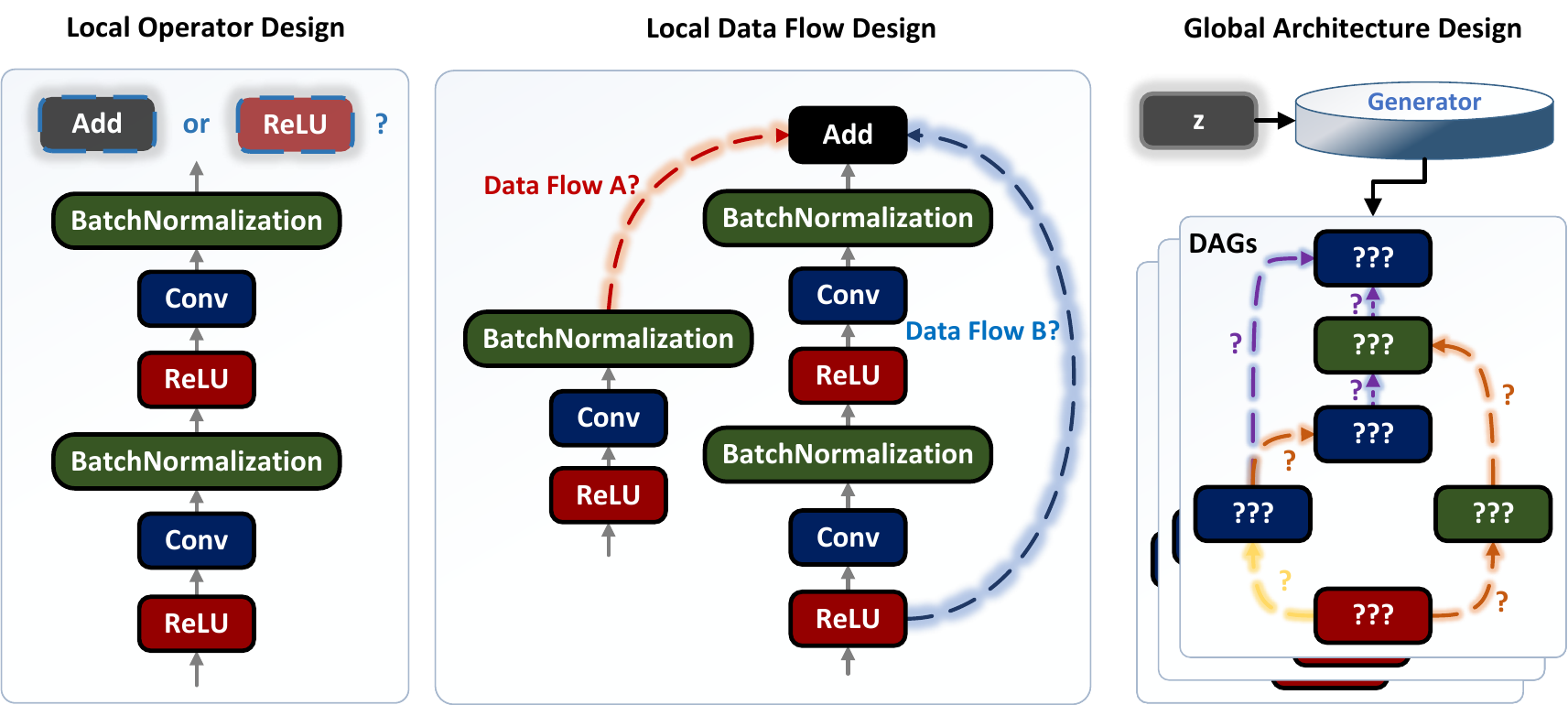}
\caption{\textbf{Paradigms of the AIGNNA}. From left to right are operator design for the local paradigm, data flow design for the local paradigm, and global architecture design for the global paradigm.}
\label{fig:all_show}
\end{figure}

To advance the development of AIGNNA based on the Younger dataset, we introduce two paradigms for neural network architecture design, each tailored to different real-world application scenarios. Figure~\ref{fig:all_show} provides an intuitive visualization of two paradigms.

\subsection{Local: Architecture Refinement In Detail}
The local paradigm addresses the need to fine-tune specific aspects of existing neural network architectures. This approach is divided into operator and data flow designs, as shown in Figure~\ref{fig:all_show}. Operator Design involves determining the most suitable type of operator for a given node based on local or global architectural information, as illustrated in the leftward of Figure~\ref{fig:all_show}. This design assesses potential replacements for current operators and suggests appropriate operators for new nodes based on neighboring structural information.

The second type, data flow design, evaluates the existence of data flows between operators. This fine-tuning method determines whether a directed edge representing data flow should connect any two nodes, utilizing insights from local and global architectural contexts.

The challenge in local paradigms arises from the vast diversity of operators and the binary nature of data flow decisions (existing or not). To assess the efficacy of this paradigm, we employ five different graph neural networks as baselines, focusing on operator and data flow design. It will be seen that operator design is more challenging than data flow design. See Section 1 for more details.

\subsection{Global: Architecture Design From Scratch}
Designing neural network architectures from scratch is an open and complex challenge. Unlike neural architecture search, which limits the search space to a predefined macro-architecture while optimizing micro-architectural elements for specific performances, the global paradigm seeks to generate comprehensive neural network architectures incorporating detailed operator-level elements from the ground up.

As shown in the rightward flowchart of Figure~\ref{fig:all_show}, this generative process is conditioned on specific properties, denoted by $z$ in Figure~\ref{fig:all_show}, such as a noise that represents the architecture's intended application or required characteristics. Moreover, the architecture's design objectives are defined by the goals it needs to achieve. Importantly, global paradigms can also iteratively leverage local paradigms to progressively achieve their comprehensive design objectives.
To explore the potential and feasibility of the global paradigm, we implement a robust baseline for validation.


\section{Experiments}\label{experiments}
We conduct two parts of experiments: one to analyze the Younger statistics information and the other to apply the paradigms.

\textbf{Homogeneous or Heterogeneous?}
Different operator types possess varying quantities and varieties of attributes. For instance, a Convolution (Conv) operator includes attributes such as dilations, kernel shape, and strides, whereas a Batch Normalization operator features attributes like epsilon and momentum. This diversity complicates the treatment of these operator types within a DAG. Consequently, it presents an exciting challenge: Should researchers treat these graphs as homogeneous or heterogeneous?

The simplest way is to treat the graphs as homogeneous, which means that all nodes are considered to be of the same type, namely `operator.'
When a graph is considered heterogeneous, all nodes are viewed as operators of different types, exhibiting heterogeneity in operator types. However, the diversity of operator types increases the complexity of analyzing and designing neural network architectures, especially from the perspective of heterogeneous graphs.

To succinctly demonstrate Younger's effectiveness in AIGNNA, we treat all architectures in Younger as homogeneous graphs to avoid introducing more variables and affecting the experimental analysis. We leave the task of treating architectures in Younger as heterogeneous graphs as future work.

\textbf{Experimental Setup:}
Since the nodes in Younger's DAGs store discrete information, such as operator types and integer attributes, it is difficult for us to process the node features of the DAGs using traditional approaches. For simplicity, we consider the operator types to be discrete features of nodes. We use two configurations: 1) Regardless of whether the operator attribute configuration is consistent, we treat the same operator type as the same type of feature, which minimizes the number of node features to the size of the ONNX operator set, denoted as `Operator w/o Attributes.' 2) For operators of the same type with different attribute configurations, they are considered as different class features, which leads to a sharp increase in the number of node features and further makes learning on the dataset more challenging, denoted as `Operator w/ Attributes.' We will see the specific effects of these two configurations in experiments.

\subsection{Statistical Analysis}
We conduct statistical analysis from two perspectives: 1) analyzes lower-dimension statistical information, such as the distribution of the number of nodes in each graph and the operator distribution in Younger. 2) analyzes high-dimension statistical information, including the distribution of three different level granularity: operator, subgraph, and graph.

\begin{figure}[!htbp]
\centering
\includegraphics[width=0.98\textwidth]{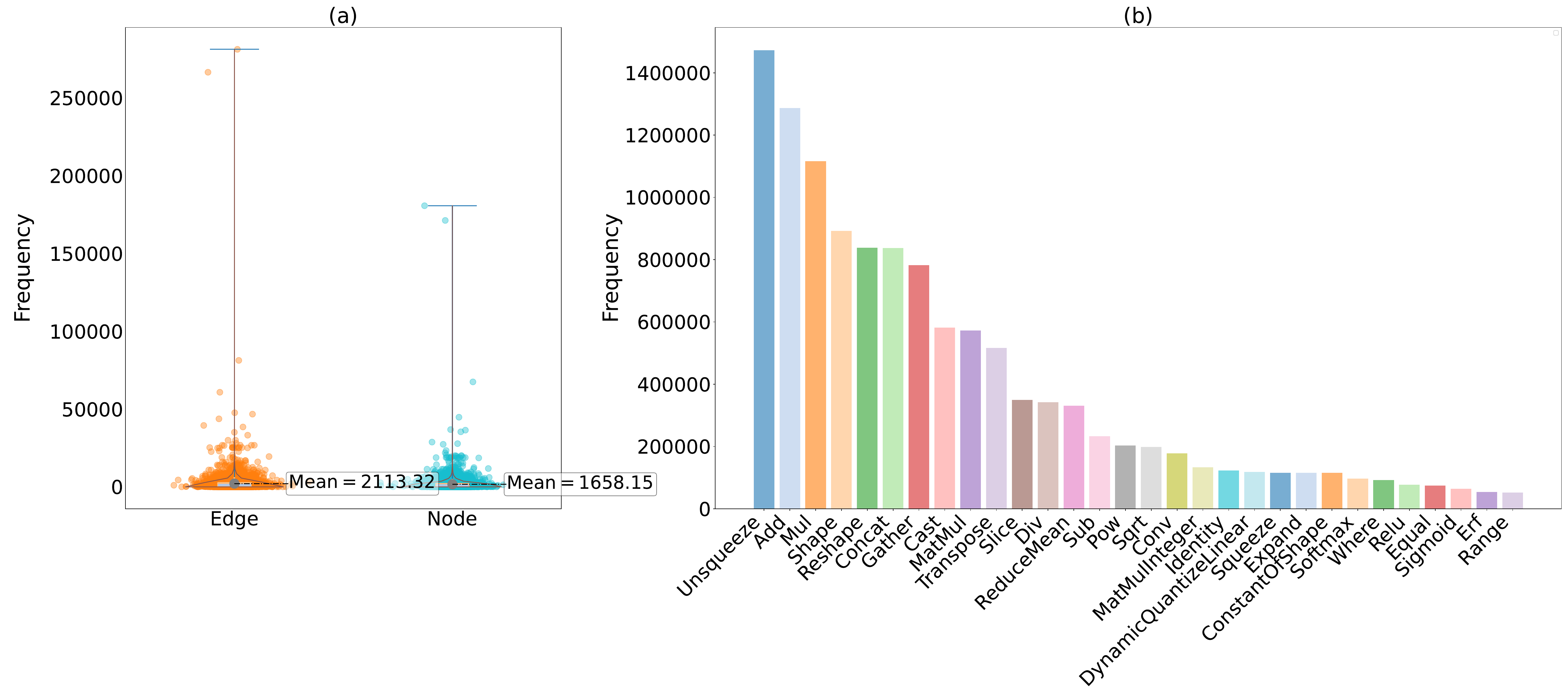}
\caption{\textbf{Distribution of \#nodes and \#edges and top 30 ONNX operators.} (a) The distribution of the number of graph nodes and edges in Younger; (b) The top 30 ONNX operators have the highest frequency in Younger.}
\label{fig:dist_n_top_30_frequency}
\end{figure}

\textbf{Low-Dimensional Statistical Information}:
We compare the statistics between Younger and conventional graph datasets. From Table~\ref{tab:dataset_gnn} and Figure~\ref{fig:dist_n_top_30_frequency}(a), Younger contains the most extensive distribution of the number of nodes in the graph, ranging from graphs containing only a dozen nodes to graphs containing hundreds of thousands of nodes. In addition, Younger also contains enough graphs compared to most graph datasets, which makes it further challenging to conduct GNNs on Younger.

Figure~\ref{fig:dist_n_top_30_frequency}(b) shows Younger's top 30 operators with the highest frequency. The dataset has a great diversity of operator types, including tensor deformations (e.g., Unsequeeze, Reshape), arithmetic operations (e.g., Add, Conv, MatMul), logical operations (e.g., Equal), and quantization (e.g., DynamicQuantizeLinear).

\begin{figure*}[!htbp]
\centering
    \begin{minipage}[t]{0.49\linewidth}
        \centering
        \includegraphics[width=\textwidth]{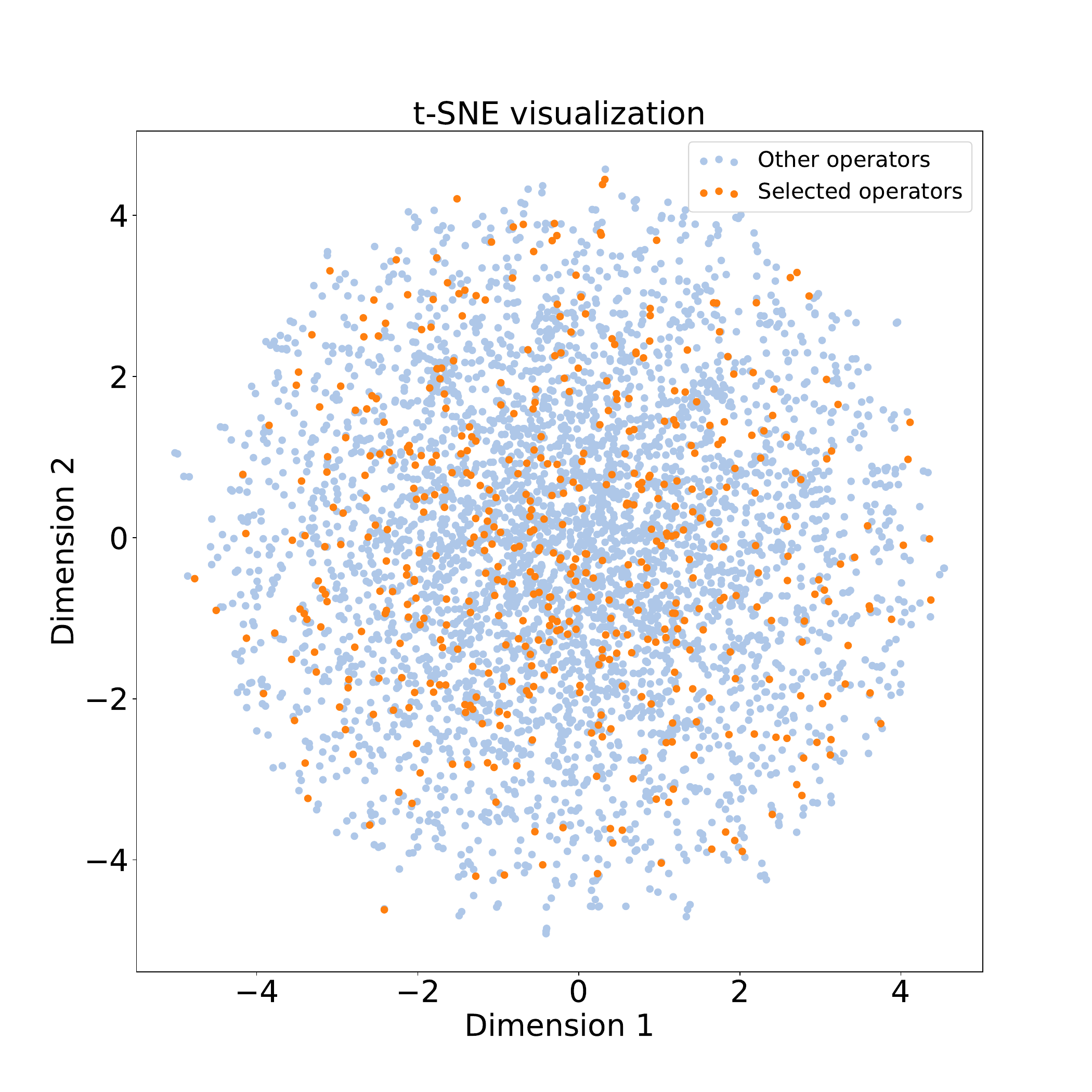}
        \vspace{-10pt}
        \captionsetup{font=scriptsize}\caption{\textbf{Node embeddings before training}}
        \label{fig:node_before}
    \end{minipage}%
    \begin{minipage}[t]{0.49\linewidth}
        \centering
        \includegraphics[width=\textwidth]{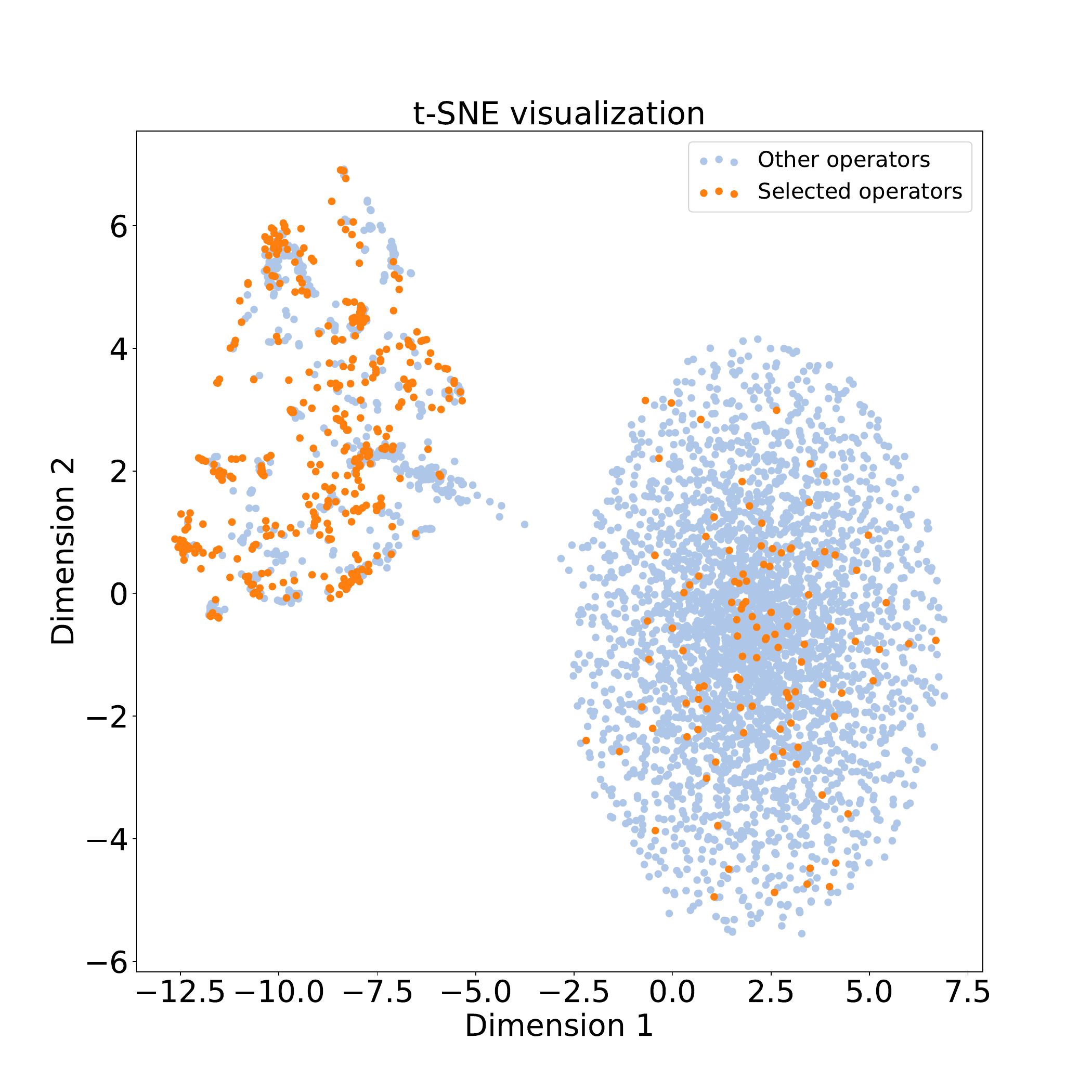}
        \vspace{-10pt}
        \captionsetup{font=scriptsize}\caption{\textbf{Node embeddings after training}}
        \label{fig:node_after}
    \end{minipage}%
    \\
    \begin{minipage}[t]{0.49\linewidth}
        \centering
        \includegraphics[width=\textwidth]{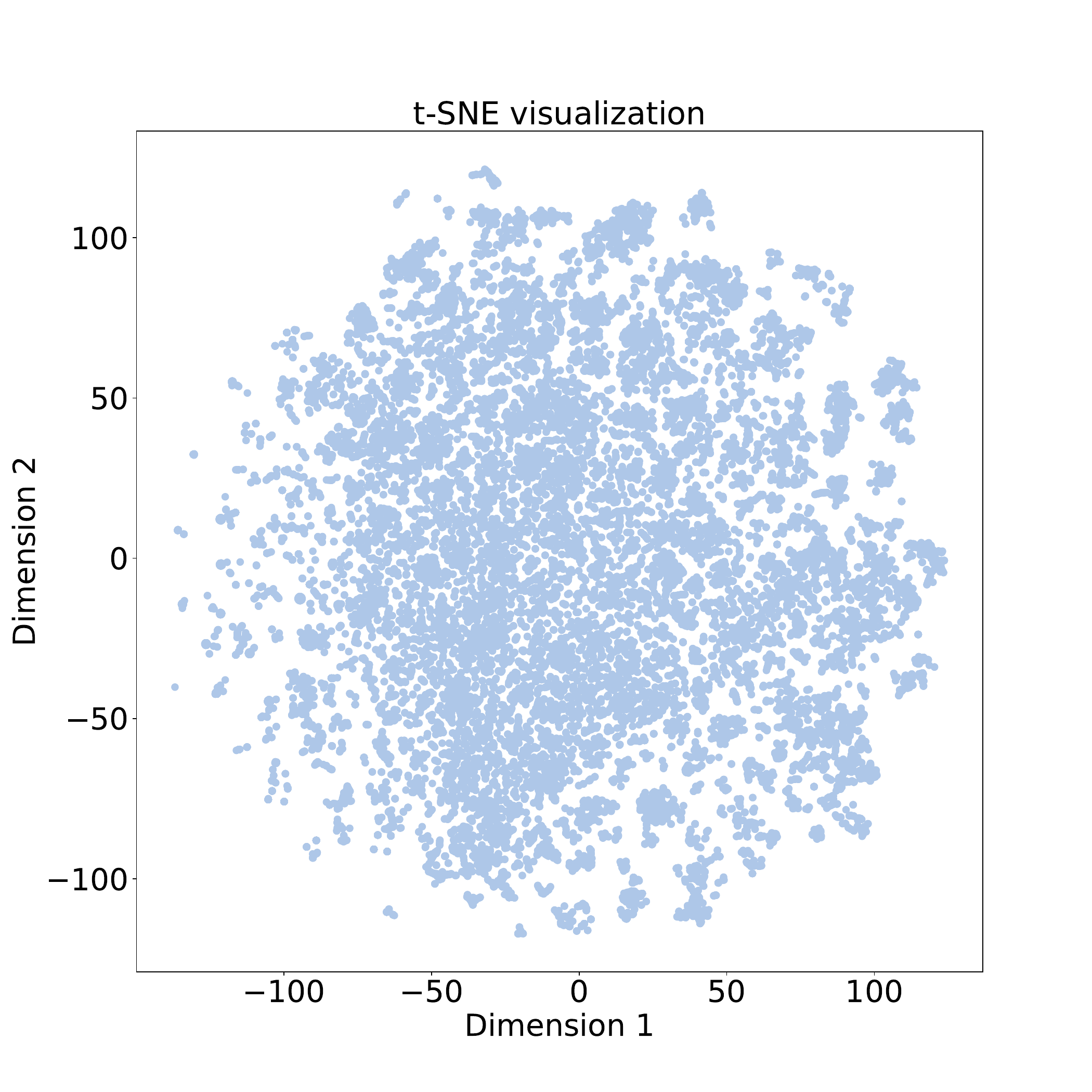}
        \vspace{-10pt}
        \captionsetup{font=scriptsize}\caption{\textbf{Subgraph embeddings}}
        \label{fig:subgraph_emb}
    \end{minipage}%
    \begin{minipage}[t]{0.49\linewidth}
        \centering
        \includegraphics[width=\textwidth]{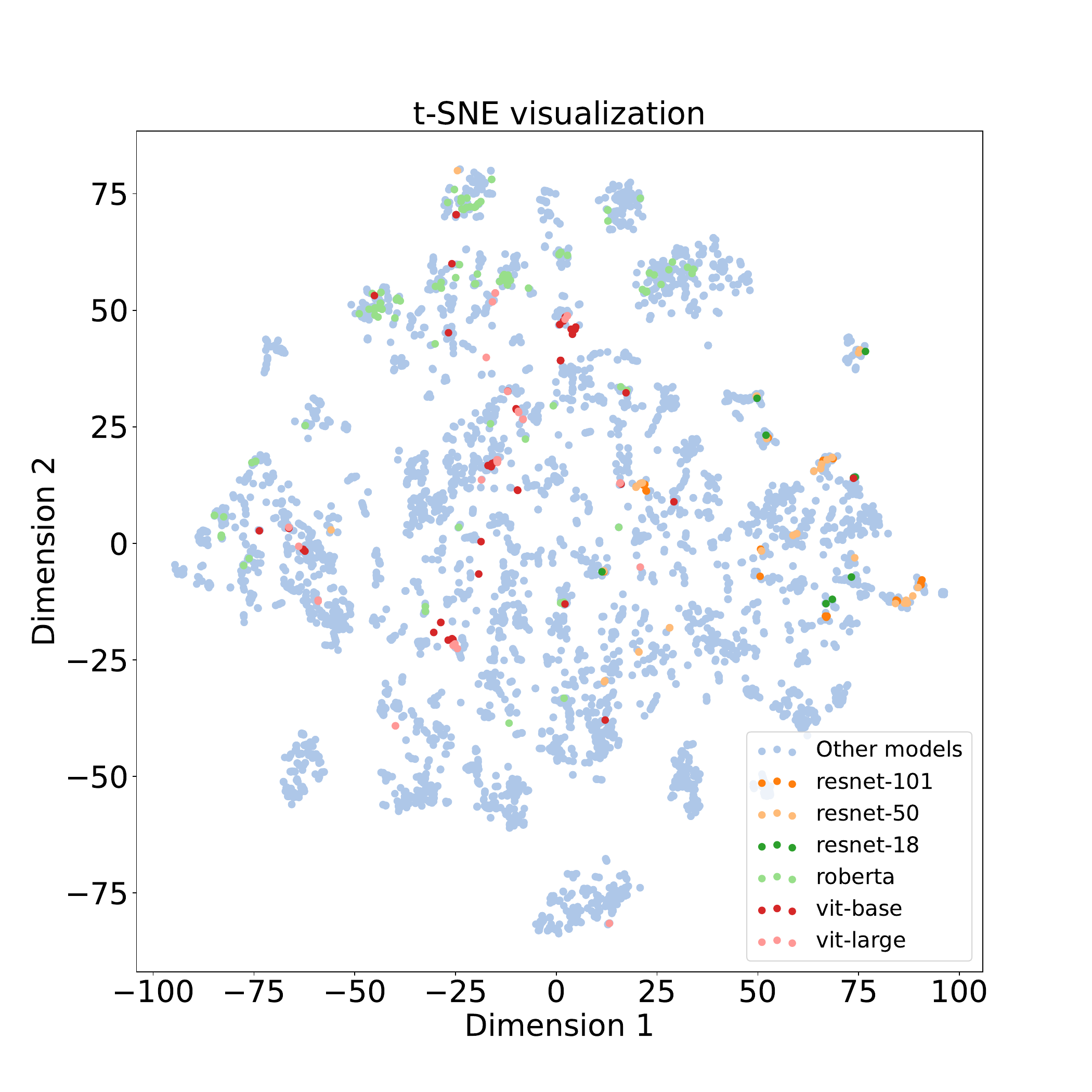}
        \vspace{-10pt}
        \captionsetup{font=scriptsize}\caption{\textbf{Graph embeddings}}
        \label{fig:graph_emb}
    \end{minipage}%
\end{figure*}

\textbf{High-Dimensional Statistical Information}
Due to the nonlinear nature of the graph, we use embedding to study the distribution properties of architectures in Younger. Specifically, the GCN network trained in subsection~\ref{local_paradigm} for operator design is used to obtain the specific embeddings.
In Figure~\ref{fig:node_before} and~\ref{fig:node_after}, orange dots represent the operators that appear in Younger's top 500 frequencies. After training, GCN gradually extracts the high-frequency operators from the original distribution and aggregates them. This reveals that learning the distribution of long-tailed operators in the dataset is a highly challenging problem; see Appendix~\ref{apx:exp} for more details.

\vspace{-10pt}

\subsection{Practical Applications}
We conducted experiments on the Younger dataset for global and local paradigms to verify the feasibility and effectiveness of our proposed two paradigms for AIGNNA. The experiment indicates that exploring AIGNNA based on Younger is feasible, and it demonstrates Younger's potential as a benchmark dataset for graph neural networks.

\subsubsection{Local Paradigm}\label{local_paradigm}
\textbf{Data Flow Design:}
We use GCN, GAT, and GraphSAGE under the data flow design paradigm to verify the effectiveness of neural architecture refinement on Younger. The results are shown in Table~\ref{tab:local_df_design}.
All models have achieved good performance on the Younger dataset, which proves that existing graph neural networks are more suitable for predicting data flows in neural network architectures.
Additionally, it can be seen that almost all models perform better without attributes because reducing the number of node features on the graph makes learning them easier.
\begin{table}[!htpb]
  \centering
\begin{tabular}{c||ccc||ccc}
\hline
\multirow{2}{*}{\textbf{Model}} &
  \multicolumn{3}{c||}{\textbf{Operator w/ Attributes}} &
  \multicolumn{3}{c}{\textbf{Operator w/o Attributes}} \\
 &
  AUC$\uparrow$ &
  F1$\uparrow$ &
  AP$\uparrow$ &
  AUC$\uparrow$ &
  F1$\uparrow$ &
  AP$\uparrow$ \\ \hline\hline
GCN~\cite{kipf2017semisupervised}       & \textbf{0.9922} & 0.7881          & \textbf{0.9913} & \textbf{0.9938} & 0.7791          & \textbf{0.9929} \\ \hline
GAT~\cite{brody2022how}       & 0.8997          & \textbf{0.8079} & 0.8720          & 0.9094          & 0.7964          & 0.8901          \\ \hline
SAGE~\cite{10.5555/3294771.3294869} & 0.9169          & 0.8033          & 0.8940          & 0.9252          & \textbf{0.8002} & 0.9026          \\ \hline
\end{tabular}
\caption{\textbf{Local paradigm: data flow design}. Bold values represent the best-performing results.}
\label{tab:local_df_design}
\end{table}%
\vspace{-10pt}

\textbf{Operator Design:}\label{operator_design}
We use five different baselines for ten experiments under the operator design paradigm, as shown in Table~\ref{tab:local_op_design}. Despite the high accuracy achieved by all baselines, the F1 score, Precision, and Recall remain low. This is primarily attributed to the complex graph structures in Younger, which are characterized by many operator types. Among these, multiple kinds of operators infrequently occur, posing challenges to achieving robust multi-classification performance. In experiments w/o attributes, we observed higher values for F1, Precision, and Recall compared to scenarios w/ attributes. This finding further substantiates the inherent complexity of the dataset and its impact on classification performance.
\begin{table}[!htpb]
\centering
\begin{tabular}{c||cccc||cccc}
\hline
\multirow{2}{*}{\textbf{Model}} & \multicolumn{4}{c||}{\textbf{Operator w/ Attributes}}            & \multicolumn{4}{c}{\textbf{Operator w/o Attributes}}                     \\
     & ACC$\uparrow$             & F1$\uparrow$ & Prec.$\uparrow$ & Recall$\uparrow$ & ACC$\uparrow$    & F1$\uparrow$ & Prec.$\uparrow$ & Recall$\uparrow$ \\ \hline
GCN~\cite{kipf2017semisupervised}  & 0.8684          & 0.1451   & 0.1713  & 0.1466  & 0.8360 & 0.2987   & 0.3657  & 0.2788  \\ \hline
GAT~\cite{brody2022how}  & OOM          & OOM   & OOM  & OOM  & 0.7139 & 0.2022   & 0.2532  & 0.2039  \\ \hline
GAE~\cite{kipf2016variational}  & \textbf{0.9016} & 0.0537   & 0.0728  & 0.0513  & 0.9073 & 0.1745   & 0.2036  & 0.1700  \\ \hline
VGAE~\cite{kipf2016variational} & 0.8243          & 0.0716   & 0.0891  & 0.0707  & 0.9137 & 0.2207   & 0.2654  & 0.2132  \\ \hline
SAGE~\cite{10.5555/3294771.3294869}              & 0.8984 & \textbf{0.2028} & \textbf{0.2383} & \textbf{0.1996} & \textbf{0.9250} & \textbf{0.3646} & \textbf{0.4323} & \textbf{0.3532} \\ \hline
\end{tabular}
\caption{\textbf{Local paradigm: operator design}. Bold values represent the best-performing results. `Prec.' in the header represents Precision.}
\label{tab:local_op_design}
\end{table}%
\vspace{-10pt}

\subsubsection{Global Paradigm}
In the global paradigm, we adopted the graph generation model DiGress, which uses a diffusion model to generate a graph. Due to the restriction of the computing resources and the maximum number of nodes in a graph of Younger reaching hundreds of thousands, we only selected architectures with node counts in the range $[1, 300]$ for training.

The DiGress we used achieved a negative log-likelihood of at least 345.4988 on the test set. Global Paradigm is a highly challenging task, so we will continue to research it in the future.


\section{Conclusion and Future Work}
This article presents Younger, a dataset of neural network architectures extracted from real-world models across various public model repositories.
Based on this dataset, we propose a new challenging field: Artificial Intelligence-Generated Neural Network Architecture (AIGNNA).
We propose two new challenges regarding neural network architecture design in this field: the Global Design Paradigm and the Local Design Paradigm. Through experiments, we have preliminarily verified the potential and effectiveness of Younger's neural architecture design in the new field. And call on more researchers to join this research direction.





\bibliographystyle{ieeetr}
\bibliography{reference}

\newpage
\begin{appendices}
\setcounter{page}{1}
\pagenumbering{roman}
\listofappendices

\section{Datasheet}

\newcommand{\edit}[1]{#1}

\subsection{Motivation}

\begin{itemize}

\item \textbf{For what purpose was the dataset created?} Was there a specific task in mind? Was there a specific gap that needed to be filled? Please provide a description.

\begin{itemize}
\item \textit{The creation of Younger aims to enable fully automated neural network architecture design and optimization, a novel domain we introduce in the paper as AIGNNA. Previous efforts in neural network architecture design and optimization have primarily relied on manually designed initial macro-architectures, followed by a search for micro-architectures within a constrained space. This approach has led to a need for more innovation and diversity in AIGNNA. The introduction of Younger addresses this deficiency. On the one hand, it enables granular (operator-level) automatic architecture design; on the other hand, as a dataset with graph structural properties, Younger can also serve as a benchmark dataset to further advance the development of graph neural networks, as demonstrated by the experiments described in the main text.}
\end{itemize}

\item \textbf{Who created the dataset (e.g., which team, research group) and on behalf of which entity (e.g., company, institution, organization)?}

\begin{itemize}
\item \textit{This dataset is presented by the author of the paper and BenchCouncil (The International Open Benchmark Council), a non-profit international organization dedicated to benchmarks and evaluations.}
\end{itemize}

\item \textbf{Who funded the creation of the dataset?} If there is an associated grant, please provide the name of the grantor and the grant name and number.

\begin{itemize}
\item \textit{Supported by the Innovation Funding of ICT, CAS under Grant No. E461070.}
\end{itemize}

\item \textbf{Any other comments?}

\begin{itemize}
\item \textit{No.}
\end{itemize}

\end{itemize}

\subsection{Composition}

\begin{itemize}

\item \textbf{What do the instances that comprise the dataset
    represent (e.g., documents, photos, people, countries)?} Are there
  multiple types of instances (e.g., movies, users, and ratings;
  people and interactions between them; nodes and edges)? Please
  provide a description.
\begin{itemize}
    \item \textit{Younger contains 7629 completely different neural network architectures (instances). Each instance is a neural network architecture from the real world, represented by a directed acyclic graph, where each node records detailed ONNX operator configurations and edges record data flow between operators. In the paper, we will discuss in detail how to view node features, that is, whether the neural network architecture should be viewed as heterogeneous or homogeneous, as detailed in Section~\ref{experiments}. The operator type is in the set of operators specified by ONNX. Currently, the latest version of ONNX is 22, supporting 193 different operators. This dataset presently includes 184 of these operators.}
\end{itemize}

\item \textbf{How many instances are there in total (of each type, if appropriate)?}
\begin{itemize}
    \item \textit{In Younger, 7629 neural network architectures were extracted from 174K real-world model species. Furthermore, Younger contains approximately 12.6M nodes and 16.1M edges.}
\end{itemize}

\item \textbf{Does the dataset contain all possible instances or is it
    a sample (not necessarily random) of instances from a larger set?}
  If the dataset is a sample, then what is the larger set? Is the
  sample representative of the larger set (e.g., geographic coverage)?
  If so, please describe how this representativeness was
  validated/verified. If it is not representative of the larger set,
  please describe why not (e.g., to cover a more diverse range of
  instances, because instances were withheld or unavailable).
\begin{itemize}
    \item \textit{As stated in Section~\ref{experiments} of the paper, the dataset was extracted from four well-known publicly available model libraries containing 743.5K models covering over 30 different task types. Due to the inability of at least half of the models to automatically convert to ONNX types, Younger's current version can only partially cover part of the real-world space. To promote the development of datasets, this work provides an online ONNX model submission platform and ONNX to DAG conversion tools to facilitate global collaborative construction.}
    \item \textit{In addition, experiments have shown that our dataset covers many mainstream architectures, so it has a certain degree of completeness.}
\end{itemize}

\item \textbf{What data does each instance consist of?} ``Raw'' data
  (e.g., unprocessed text or images) or features? In either case,
  please provide a description.
\begin{itemize}
    \item \textit{The dataset consists of three main versions. 1, the complete version includes the architecture corresponding to all unfiltered accurate models and records the source of the original model, including its model library and model ID. The model's parameter values are not recorded to ensure data security and privacy. 2. Filter versions, including all heterogeneous model architectures, to remove duplicate versions of the complete version. 3. According to specific experimental requirements, the split version has performed a certain proportion of segmentation and processing on the dataset, as described in Section~\ref{apx:exp}}
\end{itemize}

\item \textbf{Is there a label or target associated with each
    instance?} If so, please provide a description.
\begin{itemize}
    \item \textit{Each instance in the dataset contains information about operators and data flows, which can be used as labels in various AIGNNA paradigms according to the specific needs of researchers.}
\end{itemize}

\item \textbf{Is any information missing from individual instances?}
  If so, please provide a description, explaining why this information
  is missing (e.g., because it was unavailable). This does not include
  intentionally removed information, but might include, e.g., redacted
  text.
\begin{itemize}
    \item \textit{No.}
\end{itemize}

\item \textbf{Are relationships between individual instances made
    explicit (e.g., users' movie ratings, social network links)?} If
  so, please describe how these relationships are made explicit.
\begin{itemize}
    \item \textit{No.}
\end{itemize}

\item \textbf{Are there recommended data splits (e.g., training,
    development/validation, testing)?} If so, please provide a
  description of these splits, explaining the rationale behind them.
\begin{itemize}
    \item \textit{Yes. We performed dataset splitting and processing under two Local Paradigms of the AIGNNA: Local Operator Design and Local Data Flow Design. We only provide a suggested splitting and processing method, and users can make changes according to specific needs. Specifically, to verify the feasibility of the local paradigm, we processed and segmented the dataset according to different application scenarios. For Local Operator Design, we performed community detection on each architecture, removed duplicate communities, and then marked the boundary nodes of the communities as targets for predicting operator types; the communities are split into train, valid, and test. For Local Data Flow Design, we divided the entire dataset into train, valid, and test and then performed negative edge sampling on each architecture. See Section~\ref{apx:exp} for details.}
\end{itemize}

\item \textbf{Are there any errors, sources of noise, or redundancies
    in the dataset?} If so, please provide a description.
\begin{itemize}
    \item \textit{No.}
\end{itemize}

\item \textbf{Is the dataset self-contained, or does it link to or
    otherwise rely on external resources (e.g., websites, tweets,
    other datasets)?} If it links to or relies on external resources,
    a) are there guarantees that they will exist, and remain constant,
    over time; b) are there official archival versions of the complete
    dataset (i.e., including the external resources as they existed at
    the time the dataset was created); c) are there any restrictions
    (e.g., licenses, fees) associated with any of the external
    resources that might apply to a \edit{dataset consumer}? Please provide
    descriptions of all external resources and any restrictions
    associated with them, as well as links or other access points, as
    appropriate.
\begin{itemize}
    \item \textit{Younger is independent of external resources and will continue to become abundant over time. When releasing a newer dataset, we retain the older version and expand new architectures in the latest version to ensure data compatibility and persistence.}
\end{itemize}

\item \textbf{Does the dataset contain data that might be considered
    confidential (e.g., data that is protected by legal privilege or
    by doctor\edit{--}patient confidentiality, data that includes the content
    of individuals' non-public communications)?} If so, please provide
    a description.
\begin{itemize}
    \item \textit{No.}
\end{itemize}

\item \textbf{Does the dataset contain data that, if viewed directly,
    might be offensive, insulting, threatening, or might otherwise
    cause anxiety?} If so, please describe why.
\begin{itemize}
    \item \textit{No.}
\end{itemize}

\end{itemize}

\subsection{Collection Process}

\begin{itemize}

\item \textbf{How was the data associated with each instance
    acquired?} Was the data directly observable (e.g., raw text, movie
  ratings), reported by subjects (e.g., survey responses), or
  indirectly inferred/derived from other data (e.g., part-of-speech
  tags, model-based guesses for age or language)? If \edit{the} data was reported
  by subjects or indirectly inferred/derived from other data, was the
  data validated/verified? If so, please describe how.
\begin{itemize}
    \item \textit{We pull open-source models from public model hubs and convert them to ONNX format. Using self-developed tools, we extract the directed acyclic graphs contained in the ONNX format model without exposing the model parameter values while ensuring the invariance of the model architecture. Therefore, these instances cannot be directly observed, but according to the ONNX specification, our extraction tool can guarantee the accuracy of the conversion results, and manual testing is carried out through random sampling to ensure the accuracy of the data. See Section~\ref{construction} for details.}
\end{itemize}

\item \textbf{What mechanisms or procedures were used to collect the
    data (e.g., hardware apparatus\edit{es} or sensor\edit{s}, manual human
    curation, software program\edit{s}, software API\edit{s})?} How were these
    mechanisms or procedures validated?
\begin{itemize}
    \item \textit{In obtaining the original model, we called their official open-source API interfaces to integrate model information for different public model libraries and then automated the acquisition of specific models. For detailed acquisition methods, please refer to the source code instructions.}
    \item \textit{Then, we convert the models to ONNX format based on different model types. Specifically, we use the open-source Optim tool to convert the models in the Hugging Face Hub to ONNX, the open-source tf2onnx tool to convert the TensorFlow format models to ONNX, and the PyTorch built-in open-source conversion tool to convert the PyTorch format models to ONNX.}
    \item \textit{Finally, we used a self-developed DAG extraction tool to extract the DAGs contained in the ONNX model.}
    \item \textit{All the data processing work above was carried out using CPUs. We used twenty CPU servers to process the model and then manually sampled and checked the processed data to ensure the correctness of the entire process.}

\end{itemize}

\item \textbf{If the dataset is a sample from a larger set, what was
    the sampling strategy (e.g., deterministic, probabilistic with
    specific sampling probabilities)?}
\begin{itemize}
    \item \textit{The dataset was collected from multiple publicly available model libraries, and the selection of publicly available model libraries followed the model's reputation. Therefore, four well-known model libraries were selected, including Hugging Face Hub, Kaggle, ONNX Model Zoo, and PyTorch Hub.}
    \item \textit{Our collection strategy is to enumerate as many models as possible that can be converted to ONNX format to cover the entire real-world space.}
\end{itemize}

\item \textbf{Who was involved in the data collection process (e.g.,
    students, crowdworkers, contractors) and how were they compensated
    (e.g., how much were crowdworkers paid)?}
\begin{itemize}
    \item \textit{No crowdsourcing personnel participated in this work; only the authors of this paper collected and processed the data, and the work was carried out free of charge.}
\end{itemize}

\item \textbf{Over what timeframe was the data collected?} Does this
  timeframe match the creation timeframe of the data associated with
  the instances (e.g., recent crawl of old news articles)?  If not,
  please describe the timeframe in which the data associated with the
  instances was created.
\begin{itemize}
    \item \textit{All of Younger's instances were created between April and May 2024, but the time of the raw model data it extracted from may be traced back to earlier times. During Younger's creation period, the number of models in many model zoos, especially the Hugging Face Hub, is still increasing.}
\end{itemize}

\item \textbf{Were any ethical review processes conducted (e.g., by an
    institutional review board)?} If so, please provide a description
  of these review processes, including the outcomes, as well as a link
  or other access point to any supporting documentation.
\begin{itemize}
    \item \textit{No. Because Younger does not involve ethical review issues.}
\end{itemize}

\end{itemize}

\subsection{Preprocessing/cleaning/labeling}

\begin{itemize}

\item \textbf{Was any preprocessing/cleaning/labeling of the data done
    (e.g., discretization or bucketing, tokenization, part-of-speech
    tagging, SIFT feature extraction, removal of instances, processing
    of missing values)?} If so, please provide a description. If not,
  you may skip the remain\edit{ing} questions in this section.
\begin{itemize}
    \item \textit{Younger converts all operators from different versions to the latest ONNX version to ensure consistency in the architecture version.}
\end{itemize}

\item \textbf{Was the ``raw'' data saved in addition to the preprocessed/cleaned/labeled data (e.g., to support unanticipated future uses)?} If so, please provide a link or other access point to the ``raw'' data.
\begin{itemize}
    \item \textit{Younger provides the architecture of all the original models of the dataset, located at \url{https://datasets.yangs.cloud/younger/}.}
\end{itemize}

\item \textbf{Is the software \edit{that was} used to preprocess/clean/label the \edit{data} available?} If so, please provide a link or other access point.
\begin{itemize}
    \item \textit{The tool for converting different versions of operators to the same version is detailed in the source code description.}
\end{itemize}

\item \textbf{Any other comments?}
\begin{itemize}
    \item \textit{No.}
\end{itemize}

\end{itemize}

\subsection{Uses}

\begin{itemize}

\item \textbf{Has the dataset been used for any tasks already?} If so, please provide a description.
\begin{itemize}
    \item \textit{Before this paper proposed the concept of AIGNNA, Younger had not yet been applied to any task.}
\end{itemize}

\item \textbf{Is there a repository that links to any or all papers or systems that use the dataset?} If so, please provide a link or other access point.
\begin{itemize}
    \item \textit{No. However, all tasks and experiments using Younger in this article are open-source. They are at this link: \url{https://github.com/Jason-Young-AI/Younger.}}
\end{itemize}

\item \textbf{What (other) tasks could the dataset be used for?}
\begin{itemize}
    \item \textit{On the one hand, Younger supports extensive research on AIGNNA-related tasks, including but not limited to the Local and Global paradigms. The authenticity of Younger's data composition provides possibilities and new opportunities for the automated generation of neural network architectures from scratch. On the other hand, Younger's inherent graph structure properties and unique and rich statistical features naturally support graph neural network research. It can also be used as a benchmark dataset.}
\end{itemize}

\item \textbf{Is there anything about the composition of the dataset or the way it was collected and preprocessed/cleaned/labeled that might impact future uses?} For example, is there anything that a \edit{dataset consumer} might need to know to avoid uses that could result in unfair treatment of individuals or groups (e.g., stereotyping, quality of service issues) or other \edit{risks or} harms (e.g., \edit{legal risks,} financial harms\edit{)?} If so, please provide a description. Is there anything a \edit{dataset consumer} could do to mitigate these \edit{risks or} harms?
\begin{itemize}
    \item \textit{Younger excludes all model parameter values and is only related to the model architecture, so it will not bring any related issues.}
\end{itemize}

\item \textbf{Are there tasks for which the dataset should not be used?} If so, please provide a description.
\begin{itemize}
    \item \textit{No.}
\end{itemize}

\item \textbf{Any other comments?}
\begin{itemize}
    \item \textit{No.}
\end{itemize}

\end{itemize}

\subsection{Distribution}

\begin{itemize}

\item \textbf{Will the dataset be distributed to third parties outside of the entity (e.g., company, institution, organization) on behalf of which the dataset was created?} If so, please provide a description.
\begin{itemize}
    \item \textit{Yes, Younger will be open source.}
\end{itemize}

\item \textbf{How will the dataset will be distributed (e.g., tarball on website, API, GitHub)?} Does the dataset have a digital object identifier (DOI)?
\begin{itemize}
    \item \textit{We will distribute datasets for Younger at three levels, including 1. Complete version. 2. Filter version. 3. Split version. All data can be found on the official website \url{https://datasets.yangs.cloud/younger}. The retrieval methods include online browsing for on-demand downloads, tarball compression, and API retrieval. We will provide Digital Object Identifiers shortly.}
\end{itemize}

\item \textbf{When will the dataset be distributed?}
\begin{itemize}
    \item \textit{The initial version of Younger will be released before 31/06/2024 and will be continuously updated in the future to improve information related to the datasets.}
\end{itemize}

\item \textbf{Will the dataset be distributed under a copyright or other intellectual property (IP) license, and/or under applicable terms of use (ToU)?} If so, please describe this license and/or ToU, and provide a link or other access point to, or otherwise reproduce, any relevant licensing terms or ToU, as well as any fees associated with these restrictions.
\begin{itemize}
    \item \textit{\href{https://creativecommons.org/licenses/by-nc-nd/4.0/}{CC BY-NC-ND 4.0}}
\end{itemize}

\item \textbf{Have any third parties imposed IP-based or other restrictions on the data associated with the instances?} If so, please describe these restrictions, and provide a link or other access point to, or otherwise reproduce, any relevant licensing terms, as well as any fees associated with these restrictions.
\begin{itemize}
    \item \textit{The author of the paper and BenchCouncil retain ownership of the dataset and distribute it under CC BY-NC-ND 4.0.}
    \item \textit{The copyright related to the design of the model architecture in the dataset belongs to the creator of the model architecture, and BenchCouncil and the author of the paper are not responsible for this.}
\end{itemize}

\item \textbf{Do any export controls or other regulatory restrictions apply to the dataset or to individual instances?} If so, please describe these restrictions, and provide a link or other access point to, or otherwise reproduce, any supporting documentation.
\begin{itemize}
    \item \textit{No.}
\end{itemize}

\item \textbf{Any other comments?}
\begin{itemize}
    \item \textit{No.}
\end{itemize}

\end{itemize}

\subsection{Maintenance}

\begin{itemize}

\item \textbf{Who \edit{will be} supporting/hosting/maintaining the dataset?}
\begin{itemize}
    \item \textit{The authors and BenchCouncil will support, host, and maintain Younger.}
\end{itemize}

\item \textbf{How can the owner/curator/manager of the dataset be contacted (e.g., email address)?}
\begin{itemize}
    \item \textit{Please contact the first author or BenchCouncil's main office staff via email address \url{benchcouncil@gmail.com}.}
    \item \textit{Or through the website\url{ https://datasets.yangs.cloud/younger/contacts/}, contact us.}
\end{itemize}

\item \textbf{Is there an erratum?} If so, please provide a link or other access point.
\begin{itemize}
    \item \textit{Our initial version did not have an erratum, but we will release the errata in subsequent versions and publish it at the following address \url{https://datasets.yangs.cloud/younger/errata}.}
\end{itemize}

\item \textbf{Will the dataset be updated (e.g., to correct labeling
    errors, add new instances, delete instances)?} If so, please
  describe how often, by whom, and how updates will be communicated to
  \edit{dataset consumers} (e.g., mailing list, GitHub)?
\begin{itemize}
    \item \textit{Younger will be updated periodically, including adding new model architectures, removing outdated architectures, and correcting possible dataset errors. We will choose whether to update the dataset version based on whether there is a significant statistical difference between the data changes and the previous version. We hope every researcher who obtains data can fill out a nonmandatory survey questionnaire and provide a contact email. Therefore, updating notifications about Younger can be controlled through email, dataset websites, and the Younger code repository.}
\end{itemize}

\item \textbf{If the dataset relates to people, are there applicable
    limits on the retention of the data associated with the instances
    (e.g., were \edit{the} individuals in question told that their data would
    \edit{be} retained for a fixed period of time and then deleted)?} If so,
    please describe these limits and explain how they will be
    enforced.
\begin{itemize}
    \item \textit{No.}
\end{itemize}

\item \textbf{Will older versions of the dataset continue to be
    supported/hosted/maintained?} If so, please describe how. If not,
  please describe how its obsolescence will be communicated to \edit{dataset
  consumers}.
\begin{itemize}
    \item \textit{The old dataset version will continue to be supported/hosted/maintained, and all changes and maintenance of the different dataset versions will happen on the dataset's official website.}
\end{itemize}

\item \textbf{If others want to extend/augment/build on/contribute to
    the dataset, is there a mechanism for them to do so?} If so,
  please provide a description. Will these contributions be
  validated/verified? If so, please describe how. If not, why not? Is
  there a process for communicating/distributing these contributions
  to \edit{dataset consumers}? If so, please provide a description.
\begin{itemize}
    \item \textit{We support all researchers worldwide in contributing to the dataset. Still, we only support researchers submitting neural architecture models on Younger's official website to extend/augment/build on/contribute to the dataset. The neural architecture submitter will receive real-time feedback to verify whether the new architecture they submit contributes to the existing dataset.}
    \item \textit{In addition, we support researchers' private expansion of datasets to meet specific scientific research needs. Still, we do not support redistributing the modified datasets under this situation to avoid potential adverse consequences for society.}
\end{itemize}

\item \textbf{Any other comments?}
\begin{itemize}
    \item \textit{No.}
\end{itemize}

\end{itemize}

\section{Dataset Maintenance}
\subsection{Dataset Series}
Although the Younger used in this article only includes 7629 different architectures, we will still release some other series related to Younger, which are differentiated based on several critical stages of Younger's entire lifecycle to meet the various needs of researchers. The specific situation is as follows:

\subsubsection{The Complete Series}
Once all DAGs corresponding to neural network architectures are extracted from their corresponding models, they will be immediately added to the `Complete' series. Each DAG in this series of datasets will be saved in a `Dataset' class defined by us and persisted using the `Instance' class defined by us. These instances will record the public model repository to which their original models belong and their corresponding model IDs. Over time, some owners may update or delete their models. We will regularly scan existing instances and update the dataset accordingly.

\subsubsection{The Filter Series}
Once the `Complete' series is created, we will filter it by removing isomorphic DAGs (architectures), and the deduplicated results will be published as the `Filter' series. Specifically, we use the Weisfeiler Lehman (WL) Hash algorithm~\cite{shervashidzeWeisfeilerLehmanGraphKernels2011} to identify isomorphic architectures. The WL Hash algorithm iteratively hashes and aggregates the operator information of neighboring nodes of each node in the DAG to obtain the hash value of the DAG. The WL Hash algorithm ensures that isomorphic DAGs have the same hash value and strong guarantees that the hash values of non-isomorphic DAGs have significant differences. Therefore, the accuracy of the filtering results can be guaranteed.

Section~\ref{experiments} of the paper describes two configurations of operator information for nodes when using the WL hash algorithm: `Operator w/Attributes' and `Operator w/o Attributes.' Therefore, we will publish the `Filter' series separately for these two types.

\subsubsection{The Split Series}
To facilitate researchers' reproducing all the experiments in this paper and related AIGNNA research, we have provided corresponding segmented datasets for the two paradigms of AIGNNA based on the `Filter' series dataset, which we refer to as the `Split' series. For details, please look at the `Dataset Splits' subsection of Appendix~\ref{apx:exp} for each paradigm. We also publish the `Split' series separately for the `Operator w/Attributes' and `Operator w/o Attributes' types, the same as `Filter.'

\subsection{Dataset Versioning}
All versions of the series of datasets will be released follow Semantic Versioning 2.0.0~\footnote{Semantic Versioning 2.0.0~\url{https://semver.org/spec/v2.0.0.html}}. We will provide detailed version control specifications on the dataset's official website in the future, which will be briefly introduced here, given version number MAJOR MINOR.PATCH: When entries in the dataset need to be revised, the version number will increase to PATCH. When there is a significant change in the statistical characteristics of the dataset, the MINOR version will be increased. The MAJOR version number will be added when the dataset undergoes overall changes due to special reasons such as format or settings.


\section{Access and Contribute to Younger}
\subsection{Official Websites}
To involve researchers worldwide in AIGNNA research, we have established an official website~\footnote{Official Website~\url{https://datasets.yangs.cloud/younger}} for Younger, which provides functions such as querying, downloading, and contributing to datasets. The website also offers this paper's experimental code and code for dataset construction to support researchers in building private datasets. It includes the following main functions:

\subsubsection{Dataset}
Firstly, we provide the ability to select versions for downloading three different series of datasets. Secondly, we support researchers in uploading models in ONNX format, or that can be converted to ONNX format, thereby expanding Younger. Considering the issue of model privacy, we provide an offline architecture extraction tool to ensure that researchers can extract the DAG of the neural network architecture without leaking parameters and then upload the DAG through our official website. Finally, we also support online browsing, querying all neural network architectures in Younger, and generating subsets of data for on-demand selection and downloading.

\subsubsection{Code}
We open source all the source code used to build the dataset on our official website~\footnote{Source Code on Official Website~\url{https://datasets.yangs.cloud/younger/code}} and GitHub~\footnote{Source Code on GitHub~\url{https://github.com/Jason-Young-AI/Younger}}.
In addition, we also provide the source code for all the experiments involved in this paper to facilitate subsequent research, especially AIGNNA.
For information on how to use the code, please refer to the official documentation or the GitHub homepage.

\subsection{LICENSE}\label{apx:lcs}
Our dataset adopts the \href{https://creativecommons.org/licenses/by-nc-nd/4.0/}{CC BY-NC-ND 4.0} license. Before using it, please carefully read the detailed statement in the license agreement.
In addition, since our dataset is obtained from various open-source model repositories and then extracted architectures.
So, if any architecture data items violate the license declared by the model provider, please contact us, and we will cancel the publication of the relevant architecture.


\section{Experimental Details}\label{apx:exp}
This section offers a more detailed examination of the experiments discussed in the main paper. Specifically, it addresses five critical components: Local Data Flow Design and Local Operator Design within the Local Paradigm and Node, Subgraph, and Graph Embedding in the context of Statistical Analysis. It offers a comprehensive introduction and discussion of dataset splits, training details, model selection, results, and analytical insights.

\subsection{Local Data Flow Design}
\subsubsection{Dataset Splits}
Before splitting the dataset, we removed graphs with nodes or edges less than one from the `Filter' dataset. Subsequently, the dataset was divided into training, validation, and test sets in a ratio of 8:1:1 with a random seed to be set as 1234. To better meet the need for local data flow design, we removed graphs in the validation set and test set with operator type not appearing in the training set to maintain training performance. Ultimately, there were 5994, 690, and 685 unique architectures in training, validation, and test sets for node features denoted as `Operator w/ Attributes.' For node features denoted as `Operator w/o Attributes,' there were 5612, 639, and 648 unique architectures in training, validation, and test sets, respectively.

\subsubsection{Baseline Model Configuration}
\begin{figure}[!htbp]
\centering
\includegraphics[width=0.83\textwidth]{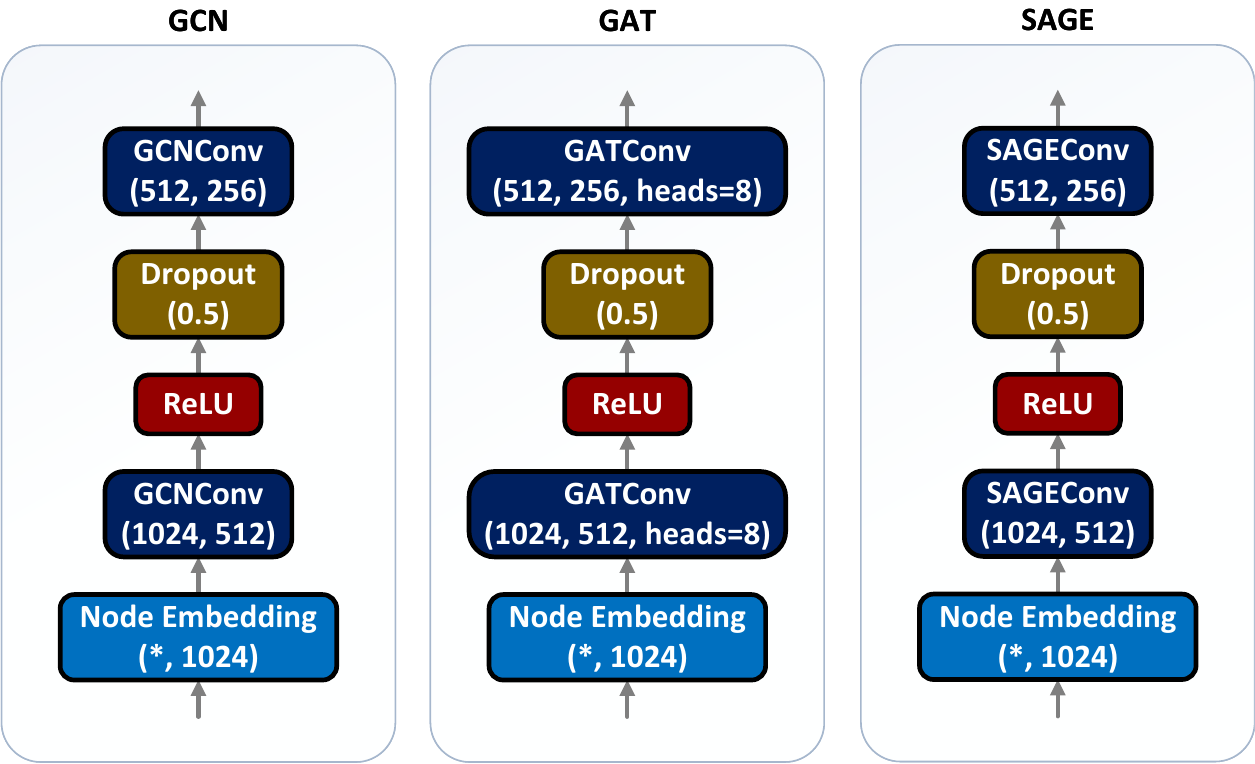}
\caption{\textbf{Topological diagram of three baseline models: GCN, GAT, and SAGE
.}}
\label{fig:df_baselines}
\end{figure}

The architectures of three baseline models represented by topological diagrams under the local data flow design paradigm are shown in Figure~\ref{fig:df_baselines} and Table~\ref{table:df_parametor} indicates the number of parameters. It is worth mentioning that the outputs of multi-head attention of GAT are averaged instead of concatenated.
\begin{table}[!htpb]
\centering
\begin{tabular}{c||c||c}
\hline
\multirow{2}{*}{\textbf{Model}} & \textbf{Operator w/ Attributes} & \textbf{Operator w/o Attributes} \\ \cline{2-3} 
                                & Number of Parameters            & Number of Parameters             \\ \hline \hline
GCN                             & 5,360,384                         & 849,664                           \\ \hline
GAT                             & 9,960,192                         & 5,449,472                          \\ \hline
SAGE                            & 6,015,744                         & 1,505,024                          \\ \hline
\end{tabular}
\caption{\textbf{Number of Parameters of Local Data Flow Design Baseline Models.}}
\label{tab:apx_local_df_parameters}
\end{table}\label{table:df_parametor}

\subsubsection{Training Configuration}\label{apx:df_training_config}
In this version, we set the random seed to 12345 and chose Adam as the optimizer for the local data flow design training process. Other hyperparameters were set as shown in Table \ref{tab:apx_train_local_df}.
The experiments for local operator design were conducted on a server running Ubuntu 22.04.1 LTS. It has four identical A800-80GB GPUs and an Intel(R) Xeon(R) Gold 6348 CPU @ 2.60GHz with 112 cores. All the baseline models for data flow design were trained on four identical A800-80GB GPUs.
\begin{table}[!htpb]
\centering
\begin{tabular}{c||ccc||ccc}
\hline
\multirow{2}{*}{\textbf{Model}} &
  \multicolumn{3}{c||}{\textbf{Operator w/ Attributes}} &
  \multicolumn{3}{c}{\textbf{Operator w/o Attributes}} \\
 &
 \quad  LR &
 \quad  WD &
  BS &
 \quad  LR &
 \quad  WD &
  BS \\ \hline\hline
GCN~\cite{kipf2017semisupervised}       &\quad 1e-4 & \quad 5e-5         & 1 &\quad  1e-4 &\quad  5e-5          & 1 \\ \hline
GAT~\cite{brody2022how}       & \quad 1e-4 &\quad  5e-5         & 1 & \quad 1e-4 &\quad  5e-5          & 1          \\ \hline
SAGE~\cite{10.5555/3294771.3294869} &\quad  1e-4 & \quad 5e-5         & 1 & \quad 1e-4 &\quad  5e-5          & 1          \\ \hline
\end{tabular}
\caption{\textbf{Training Details of Local Data Flow Design}.  `LR,' `WD,' and `BS' in the header represent Learning Rate, Weight Decay, and Batch Size, respectively}
\label{tab:apx_train_local_df}
\end{table}

\subsubsection{Metrics}
\textbf{Area under the Receiver Operating Characteristic Curve (AUC)}:
\begin{equation}
\text{TPR} = \frac{\text{TP}}{\text{TP} + \text{FN}},
\end{equation}
\begin{equation}
\text{FPR} = \frac{\text{FP}}{\text{FP} + \text{TN}}.
\end{equation}
For the Receiver Operating Characteristic (ROC) Curve, the Y axis represents the true positive rate ($\text{TPR}$) while the X axis represents the false positive rate ($\text{FPR}$).
A value of AUC close to 1 represents a better classification prediction performance.

\textbf{F1 Score (F1)}:
\begin{equation}
\text{F1} = \frac{2 \cdot \text{TP}}{2 \cdot \text{TP} + \text{FP} + \text{FN}},
\end{equation}
where $\text{TP}$, $\text{FP}$, and $\text{FN}$ represent the number of true positives, false positives, and false negatives, respectively.

\textbf{Average Precision (AP)}:
\begin{equation}
\text{AP} = \sum_{n=1}^{N} (R_n - R_{n-1}) P_n,
\end{equation}
where $R$ and $P$ represent the precision and recall, while $n$ denotes the $n$th threshold.

\subsubsection{Checkpoint Selection}
We chose checkpoints to test the performance of baseline models based on the weighted average of all the metrics reported during validation. The weighted averages of AUC, F1, and AP were calculated to measure the performance of baseline models. In this version, all weights are set to be the same.

\subsubsection{Results and Analysis}
We set up our configuration as stated in Section~\ref{apx:df_training_config} and used GCN, GAT, and GraphSAGE for six experiments under the data flow design paradigm on Younger. As shown in Table~\ref{tab:apx_local_df}, these three baseline models perform well on all metrics. It is worth noting that GCN outperforms other models in all metrics except F1 Score, regardless of whether the operators have attributes. 
\begin{table}[!htpb]
  \centering
\begin{tabular}{c||ccc||ccc}
\hline
\multirow{2}{*}{\textbf{Model}} &
  \multicolumn{3}{c||}{\textbf{Operator w/ Attributes}} &
  \multicolumn{3}{c}{\textbf{Operator w/o Attributes}} \\
 &
  AUC$\uparrow$ &
  F1$\uparrow$ &
  AP$\uparrow$ &
  AUC$\uparrow$ &
  F1$\uparrow$ &
  AP$\uparrow$ \\ \hline\hline
GCN~\cite{kipf2017semisupervised}       & \textbf{0.9933}& 0.7893& \textbf{0.9924}& \textbf{0.9949}& 0.7907& \textbf{0.9942}\\ \hline
GAT~\cite{brody2022how}       & 0.9195& \textbf{0.8023}& 0.8974& 0.9133& 0.7960& 0.8937\\ \hline
SAGE~\cite{10.5555/3294771.3294869} & 0.9702& 0.8005& 0.9682& 0.8991& \textbf{0.8053}& 0.8591\\ \hline
\end{tabular}
\caption{\textbf{Local paradigm: data flow design}. Bold values represent the best-performing results.}
\label{tab:apx_local_df}
\end{table}%

\subsection{Local Operator Design}
\subsubsection{Dataset Splits}
Due to the lack of relevant research on extracting building blocks for neural network architecture. Therefore, we performed community detection on all DAGs (Neural Network Architecture) in the `Filter' dataset to extract the building blocks of the neural network architecture. Through community detection, we can identify the closely connected node sets in the graph to help identify subsets of nodes with specific correlations or functional associations. Although there is no evidence to suggest that the subgraphs extracted by community detection are effective building blocks for neural network architecture, in this paper, it is reasonable to use this method to extract subgraphs for preliminary validation to test the feasibility of Local Operator Design.

We adopt the Clauset Newman Moore Grey modularity maximization method~\cite{PhysRevE.70.066111} as the community detection algorithm and set it to detect at least one community, the DAG itself.
For each community, we simultaneously query its node boundary and label it as the node to be predicted. The community and node boundary form a new subgraph, and the definition of node boundary is shown in Formula~\ref{eq:nodeb}.
\begin{equation}
    \mathcal{B} = \{v | v \in \mathcal{D} - \mathcal{C}, u \in \mathcal{C}, (u, v) \in \mathcal{E}\},
\end{equation}\label{eq:nodeb}
where $\mathcal{D}$, $\mathcal{C}$, and $\mathcal{E}$ represent the node set of DAG and the node set of community and edge set of DAG, respectively, and $(u, v)$ indicates two directed edges $<u, v>$ and $<v, u>$.

Finally, we will deduplicate the subgraphs formed by all community and node boundary pairs, i.e., remove isomorphic subgraphs. Finally, 38,803 and 29,581 non-isomorphic subgraphs were obtained under the configurations of `Operator w/ Attributes' and `Operator w/o Attributes', respectively.
To obtain the final training, validation, and test sets, we split all non-isomorphic subgraphs in an 8:1:1 ratio. Specifically, under the `Operator w/ Attribute' configuration, the training, validation, and testing sets contain 31,282, 3,769, and 3,752 subgraphs, respectively, while under the `Operator w/o Attribute' configuration, they include 23,775, 2,907 and 2,899 subgraphs, respectively.

\subsubsection{Baseline Model Configuration}
\begin{figure}[!htbp]
\centering
\includegraphics[width=0.83\textwidth]{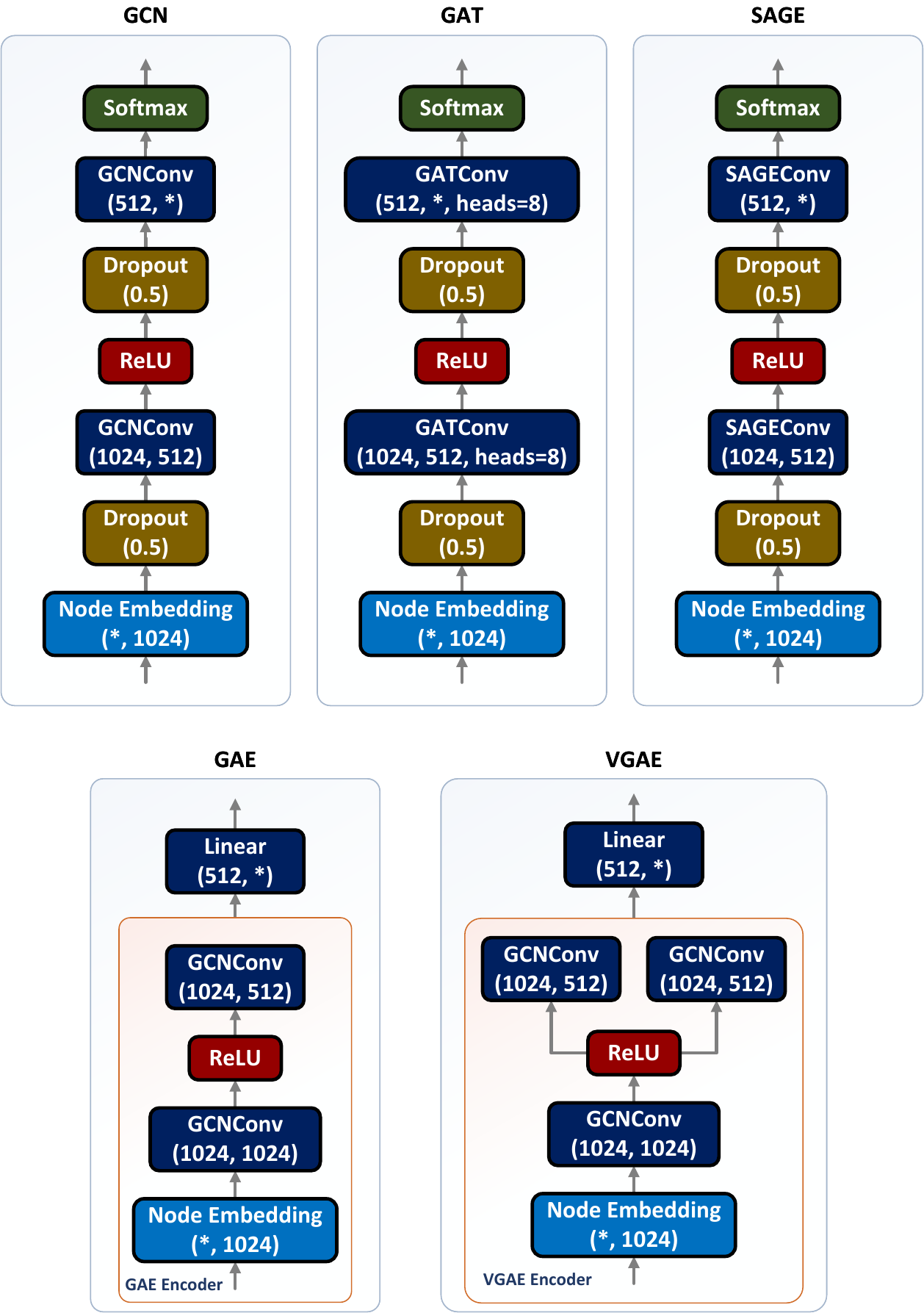}
\caption{\textbf{Topological diagram of five baseline models: GCN, GAT, SAGE, GAE, and VGAE
.}}
\label{fig:op_baselines}
\end{figure}

The architectures of baseline models represented by topological diagrams under the local operator design paradigm are shown in Figure~\ref{fig:op_baselines} and Table~\ref{table:op_parameters} indicates the number of parameters.
For experiments with GAE and VGAE under the local operator design paradigm, we first pre-trained the encoders of GAE and VGAE, then trained the linear layers for classification using the output from encoders. For GAT, the outputs of multi-head attention of GAT are averaged instead of concatenated.
\begin{table}[!htpb]
\centering
\begin{tabular}{c||c||c}
\hline
\multirow{2}{*}{\textbf{Model}} & \textbf{Operator w/ Attributes} & \textbf{Operator w/o Attributes} \\ \cline{2-3} 
                                & Number of Parameters            & Number of Parameters             \\ \hline \hline
GCN                             & 7,301,433                       & 809,145                          \\ \hline
GAT                             & 26,852,041                      & 5,153,353                        \\ \hline
SAGE                            & 10,083,129                      & 1,428,153                        \\ \hline
GAE-Encoder                     & 6,089,216                       & 1,763,840                        \\ \hline
GAE-Classification              & 2,261,817                       & 94,905                           \\ \hline
VGAE-Encoder                    & 6,614,016                       & 2,288,640                        \\ \hline
VGAE-Classification             & 2,261,817                       & 94,905                           \\ \hline
\end{tabular}
\caption{\textbf{Number of Parameters of Local Operator Design Baseline Models.}}
\label{tab:apx_local_op_parameters}
\end{table}\label{table:op_parameters}

\vspace{-16pt}

\subsubsection{Training Configuration}\label{apx:op_training_config}
In this version, we set the random seed to 12345 and chose Adam as the optimizer for the local operator design training process. Other hyperparameters were set as shown in the Table~\ref{tab:apx_train_local_op}.
The experiments for local operator design were conducted on a server running Ubuntu 22.04.1 LTS. It has four identical A800-80GB GPUs and an Intel(R) Xeon(R) Gold 6348 CPU @ 2.60GHz with 112 cores. GAT, GCN, and SAGE were trained on four A800-80GB GPUs, while GAE and VGAE were trained on one A800-80GB GPU. 
\begin{table}[!htpb]
\centering
\begin{tabular}{c||ccc||ccc}
\hline
\multirow{2}{*}{\textbf{Model}} & \multicolumn{3}{c||}{\textbf{Operator w/ Attributes}} & \multicolumn{3}{c}{\textbf{Operator w/o Attributes}} \\
                                & LR     & WD     & BS    & LR     &  \quad WD     & BS    \\ \hline\hline
GCN~\cite{kipf2017semisupervised}                             & 1e-3              & 5e-5             & 512           & 1e-3              & \quad  5e-5             & 512           \\ \hline
GAT~\cite{brody2022how}                             & OOM               & OOM              & OOM           & 1e-3              & \quad  5e-5             & 512           \\ \hline
SAGE~\cite{10.5555/3294771.3294869}                             & 1e-3              & 5e-5             & 512           & 1e-3              & \quad  5e-5             & 512           \\ \hline
GAE-Encoder~\cite{kipf2016variational}                     & 1e-4              & 5e-5             & 512           & 1e-4              & \quad  5e-5             & 512           \\ \hline
GAE-Classification~\cite{kipf2016variational}                      & 1e-3              & 5e-4             & 512           & 1e-3              &  \quad 5e-4             & 512           \\ \hline
VGAE-Encoder~\cite{kipf2016variational}                    & 1e-4              & 5e-5             & 512           & 1e-4              & \quad  5e-5             & 512           \\ \hline
VGAE-Classification~\cite{kipf2016variational}                     & 1e-3              & 5e-4             & 512           & 1e-3              & \quad 5e-4             & 512           \\ \hline
\end{tabular}
\caption{\textbf{Training Details of Local Operator Design}. `LR,' `WD,' and `BS' in the header represent Learning Rate, Weight Decay, and Batch Size, respectively.}
\label{tab:apx_train_local_op}
\end{table}

\subsubsection{Metrics}
\textbf{Accuracy (ACC)}: The ratio of correctly predicted instances to the total instances.

\textbf{F1 Score (F1)}:
\begin{equation}
\text{F1} = \frac{2 \cdot \text{TP}}{2 \cdot \text{TP} + \text{FP} + \text{FN}},
\end{equation}
where $\text{TP}$, $\text{FP}$, and $\text{FN}$ represent the number of true positives, false positives, and false negatives, respectively.

\textbf{Precision (Prec)}:
\begin{equation}
\text{Precision} = \frac{\text{TP}}{\text{TP} + \text{FP}},
\end{equation}
where $\text{TP}$ and $\text{FP}$ represent the number of true positives and false positives.

\textbf{Recall}:
\begin{equation}
\text{Recall} = \frac{\text{TP}}{\text{TP} + \text{FN}},
\end{equation}
where $\text{TP}$ and $\text{FN}$ represent the number of true positives and false negatives.

\subsubsection{Checkpoint Selection}\label{apx:df_design_checkpoint}
We chose checkpoints to test the performance of baseline models based on the weighted average of ACC, F1 Score, Precision, and Recall reported during validation. In this version, all weights are set to be the same. For the encoder of GAE and VGAE, we chose the checkpoint on training step 4000, whose training loss remained stable. 

\subsubsection{Results and Analysis}
We set configuration as stated in Section \ref{apx:op_training_config}. Baseline models, including GCN, GAT, GAE, VGAE, and SAGE, were used under the operator design paradigm. As shown in Table~\ref{tab:apx_local_op}, all baseline models achieve high accuracy but perform poorly in other metrics. The reason can be attributed to the complexity of Younger and further to the complexity of the neural network architectures in the real world. Another reason is that some typical types of operators appear more frequently while others appear less frequently, causing the model to be biased toward predicting the majority of operators. It can be seen that all baseline models in experiments w/o attributes achieve higher F1, Precision, and Recall compared to those in experiments w/ attributes. This indicates that reducing the variety of operators and making their distribution more uniform can improve the multi-classification performance. In addition, among these baseline models, SAGE performs excellently on almost all metrics. Notice that GAT lacks experiments with Operator w/ Attributes due to excessively large parameter counts as shown in Table \ref{tab:apx_local_op_parameters}, resulting in out-of-memory issues during execution.
\begin{table}[!htpb]
\centering
\begin{tabular}{c||cccc||cccc}
\hline
\multirow{2}{*}{\textbf{Model}} & \multicolumn{4}{c||}{\textbf{Operator w/ Attributes}}            & \multicolumn{4}{c}{\textbf{Operator w/o Attributes}}                     \\
     & ACC$\uparrow$             & F1$\uparrow$ & Prec.$\uparrow$ & Recall$\uparrow$ & ACC$\uparrow$    & F1$\uparrow$ & Prec.$\uparrow$ & Recall$\uparrow$ \\ \hline
GCN~\cite{kipf2017semisupervised}  & 0.7454& 0.1294& 0.1666& 0.1323& 0.7627& 0.2988& 0.3750& 0.2941\\ \hline
GAT~\cite{brody2022how}  & OOM          & OOM   & OOM  & OOM  & 0.7163& 0.2007& 0.2519& 0.1979\\ \hline
GAE~\cite{kipf2016variational}  & 0.8173& 0.0484& 0.0658& 0.0467& 0.8179& 0.1514& 0.1815& 0.1438\\ \hline
VGAE~\cite{kipf2016variational} & \textbf{0.8224}& 0.0724& 0.0924& 0.0712& 0.8243& 0.1969& 0.2500& 0.1881\\ \hline
SAGE~\cite{10.5555/3294771.3294869}              & 0.8049& \textbf{0.1927}& \textbf{0.2385}& \textbf{0.1878}& \textbf{0.9238}& \textbf{0.3477}& \textbf{0.4144}& \textbf{0.3375}\\ \hline
\end{tabular}
\caption{\textbf{Local paradigm: operator design}. Bold values represent the best-performing results. `Prec.' in the header represents Precision.}
\label{tab:apx_local_op}
\end{table}%

\subsection{Node Embedding}
\subsubsection{Checkpoint Selection}
To better illustrate the distribution of operators in Younger in high-dimensional space, we selected checkpoints of baseline models according to the method from section \ref{apx:df_design_checkpoint} and then extracted the embeddings of operators with attributes and those without attributes from node embedding layer of baseline models. Due to the problem about memory overflow, the visualization of `Operator w/o Attributes' about GAT is not presented. To compare the training effectiveness, we also extracted the embeddings from the initial node embedding layer without loading any checkpoints.
\subsubsection{Visualization}
\begin{figure*}[!htbp]
\centering
    \begin{minipage}[t]{0.45\linewidth}
        \centering
        \includegraphics[width=\textwidth]{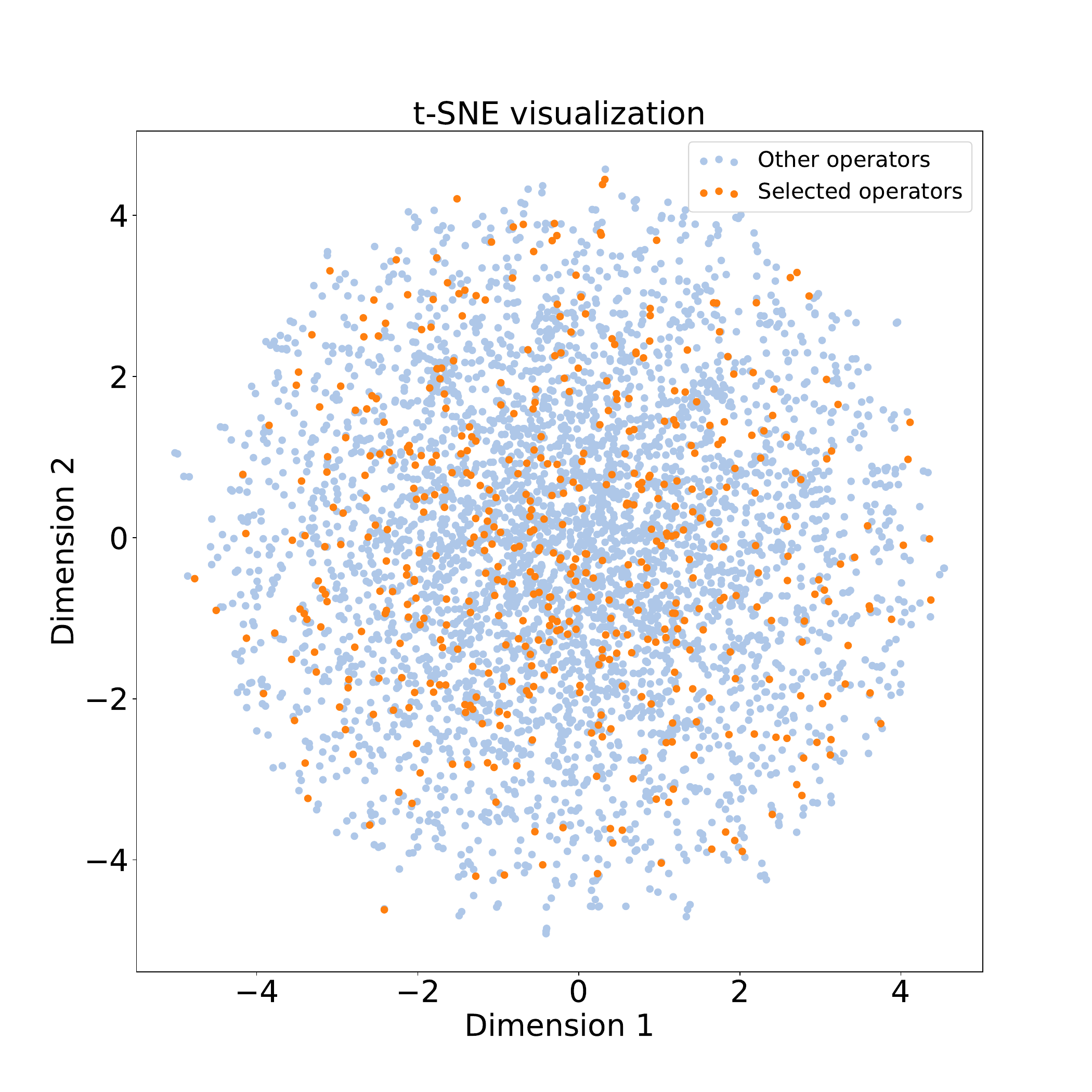}
        \vspace{-10pt}
        \captionsetup{font=scriptsize,justification=centering}\caption{\textbf{Node embeddings before training \\ (GCN - Operator w/ Attributes)}}
        \label{fig:gcn_node_before}
    \end{minipage}%
    \begin{minipage}[t]{0.45\linewidth}
        \centering
        \includegraphics[width=\textwidth]{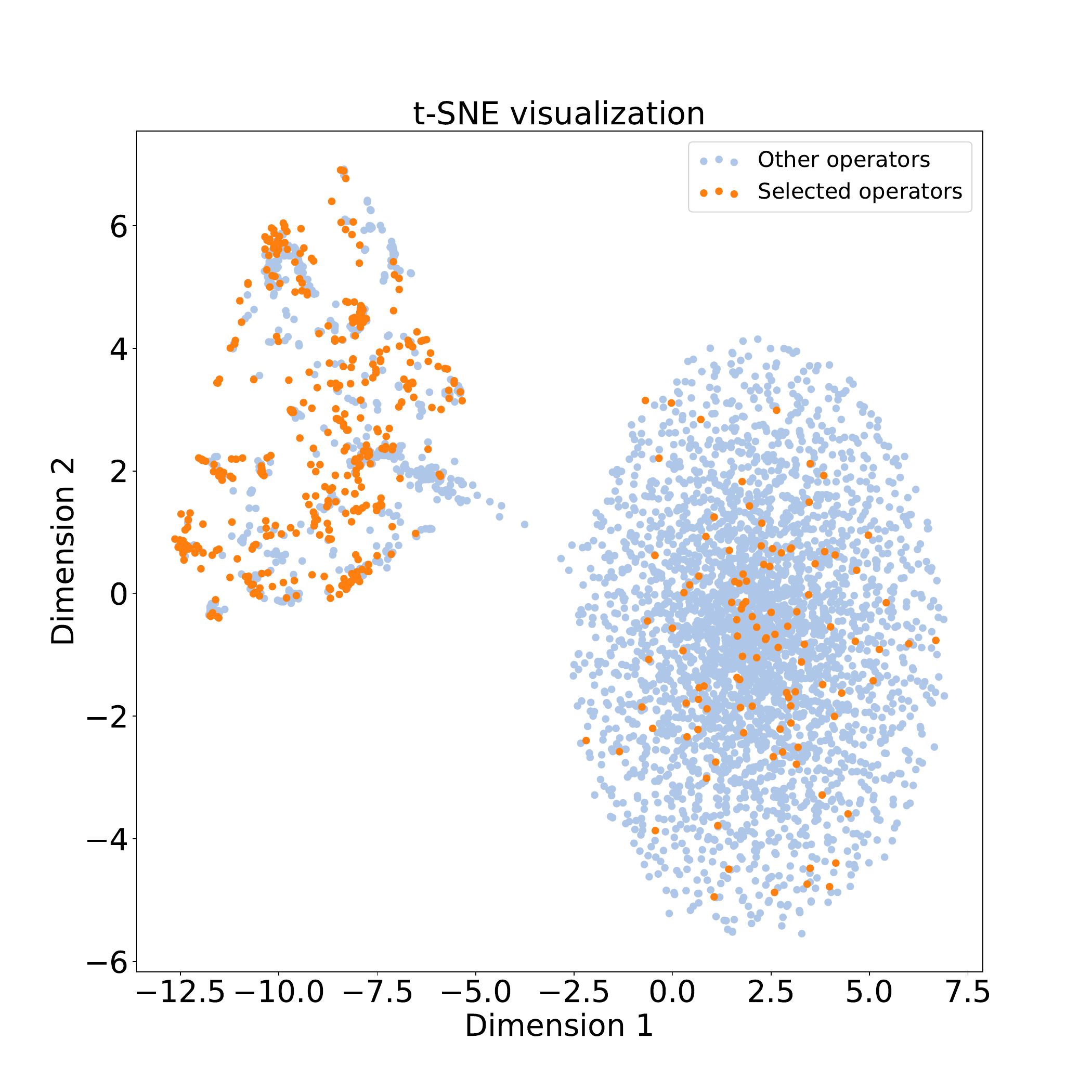}
        \vspace{-10pt}
        \captionsetup{font=scriptsize,justification=centering}\caption{\textbf{Node embeddings after training \\ (GCN - Operator w/ Attributes)}}
        \label{fig:gcn_node_after}
    \end{minipage}%
\end{figure*}

\begin{figure*}[!htbp]
\centering
    \begin{minipage}[t]{0.45\linewidth}
        \centering
        \includegraphics[width=\textwidth]{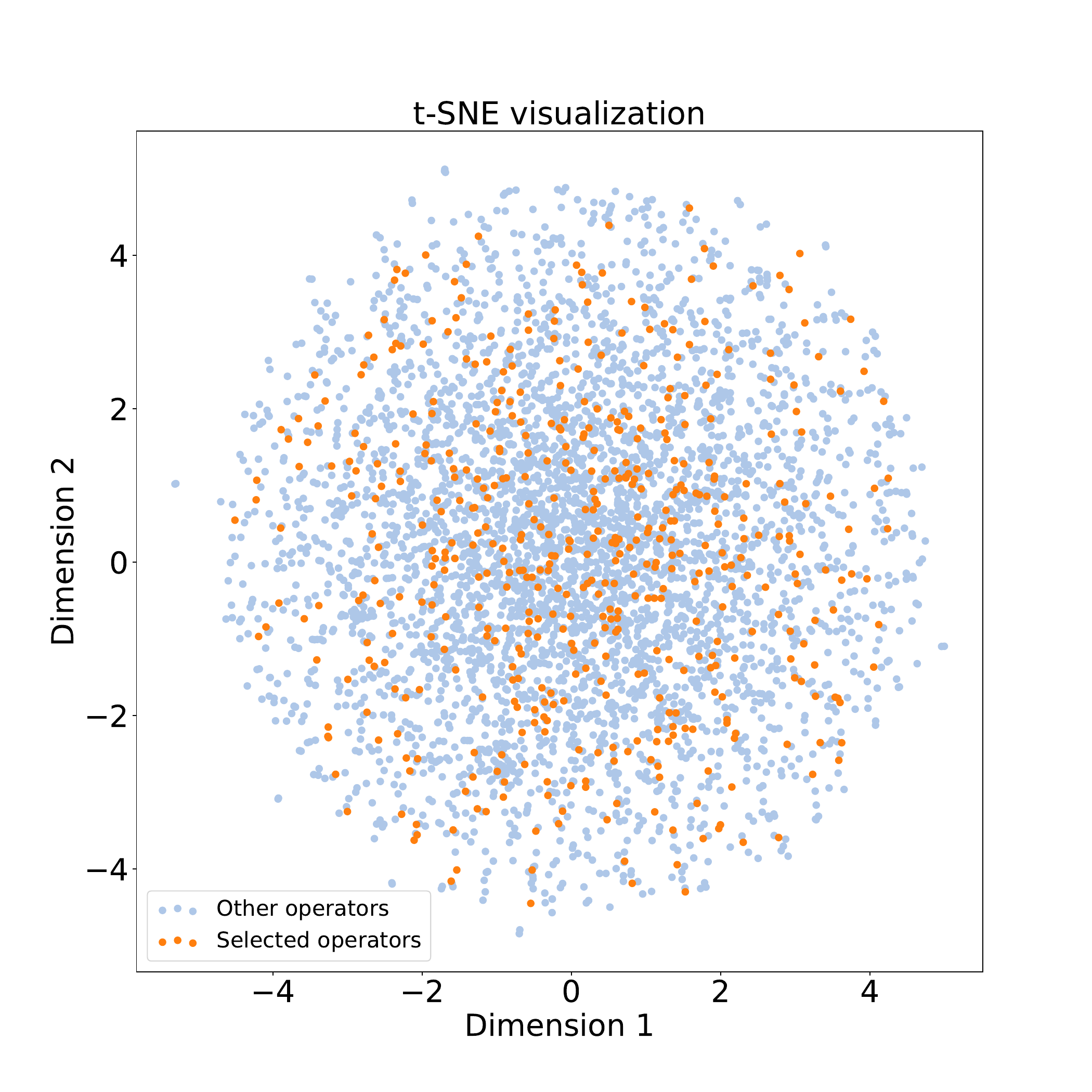}
        \vspace{-10pt}
        \captionsetup{font=scriptsize,justification=centering}\caption{\textbf{Node embeddings before training \\ (SAGE - Operator w/ Attributes)}}
        \label{fig:sage_node_before}
    \end{minipage}%
    \begin{minipage}[t]{0.45\linewidth}
        \centering
        \includegraphics[width=\textwidth]{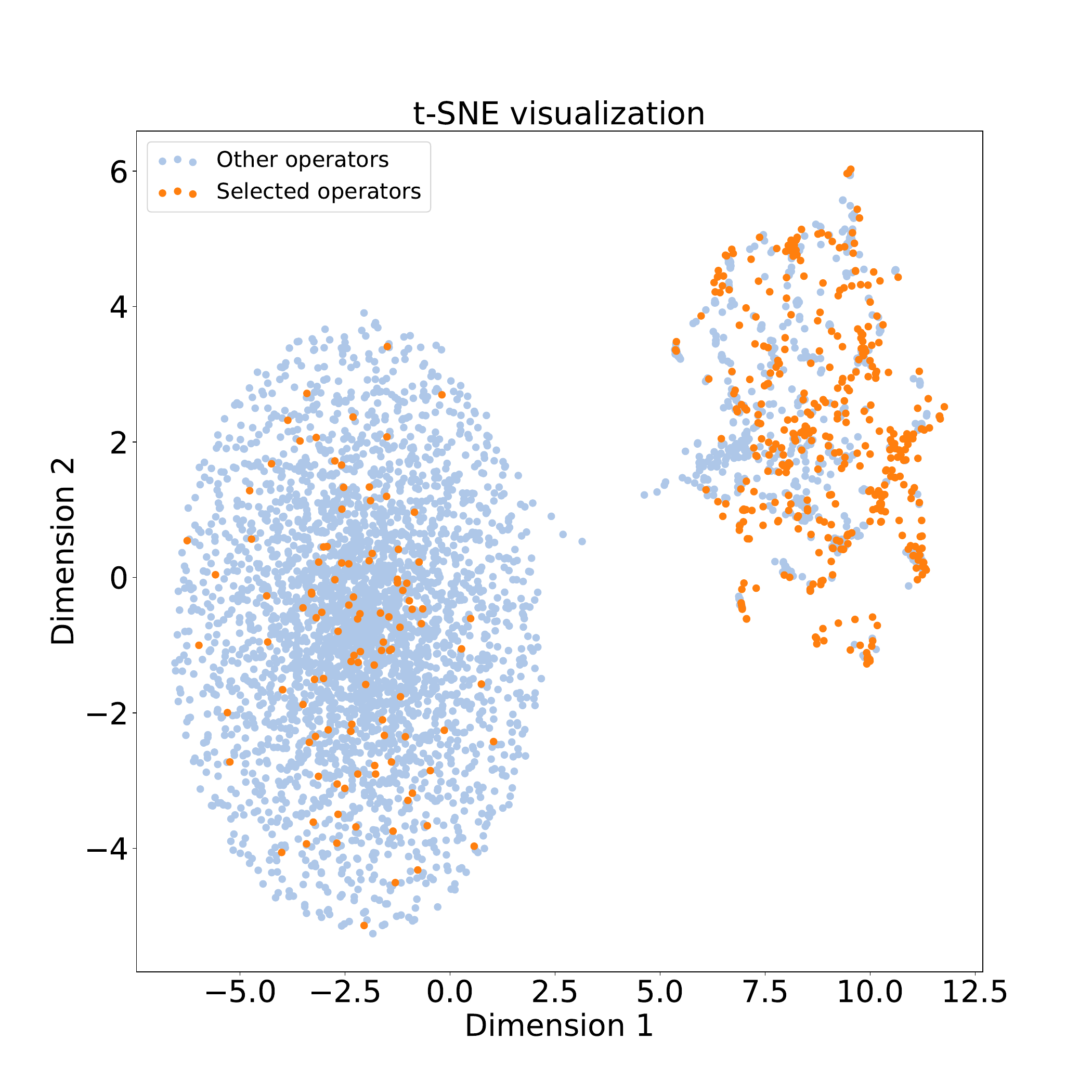}
        \vspace{-10pt}
        \captionsetup{font=scriptsize,justification=centering}\caption{\textbf{Node embeddings after training \\ (SAGE - Operator w/ Attributes)}}
        \label{fig:sage_node_after}
    \end{minipage}%
\end{figure*}

Figure \ref{fig:gcn_node_before}-\ref{fig:sage_node_after} show the t-SNE visualization results of node embeddings before and after training from GCN and SAGE with node features denoted as `Operator w/ Attributes.' The orange points represent Younger's top 500 most frequently occurring operators. It can be observed that before training, the distribution of node embeddings is relatively concentrated and chaotic. After training, the distribution of embeddings representing high-frequency nodes selected and other low-frequency nodes from Younger was well distinguished. This indicates an uneven distribution of node quantities among different types, which introduces bias into the learning process of baseline models.

\begin{figure*}[!htbp]
\centering
    \begin{minipage}[t]{0.45\linewidth}
        \centering
        \includegraphics[width=\textwidth]{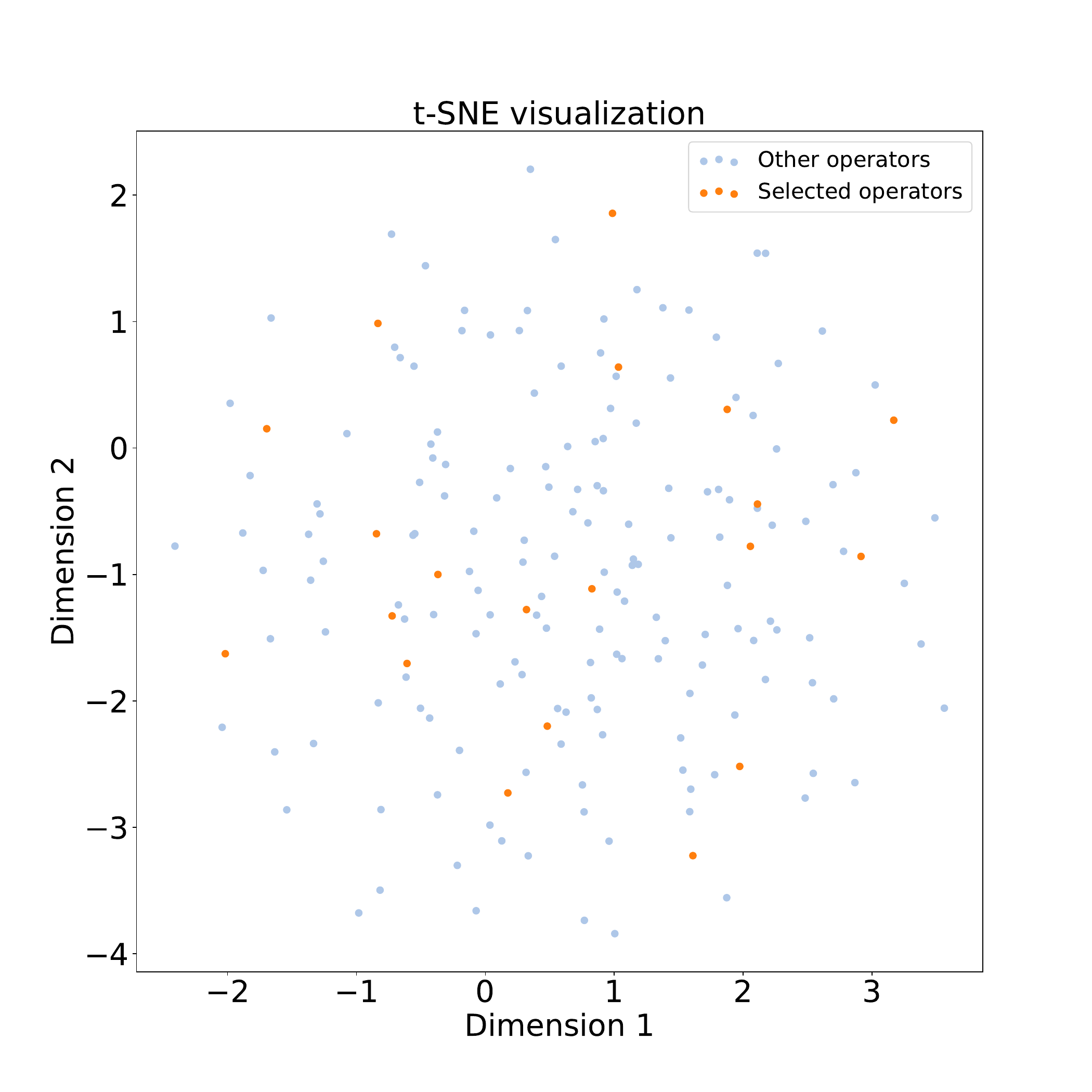}
        \vspace{-10pt}
        \captionsetup{font=scriptsize,justification=centering}\caption{\textbf{Node embeddings before training \\ (GCN - Operator w/o Attributes)}}
        \label{fig:gcn_operator_before}
    \end{minipage}%
    \begin{minipage}[t]{0.45\linewidth}
        \centering
        \includegraphics[width=\textwidth]{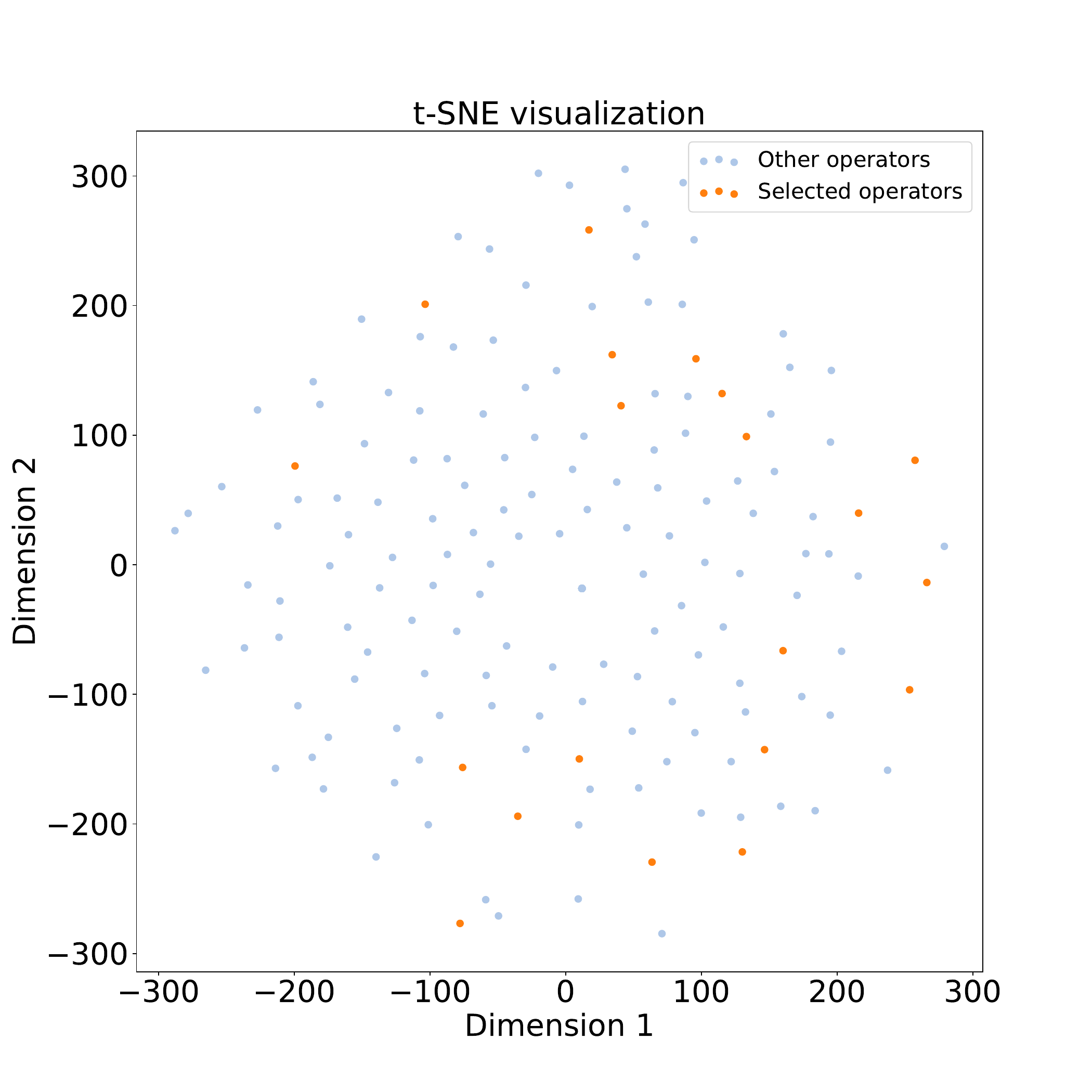}
        \vspace{-10pt}
        \captionsetup{font=scriptsize,justification=centering}\caption{\textbf{Node embeddings after training \\ (GCN - Operator w/o Attributes)}}
        \label{fig:gcn_operator_after}
    \end{minipage}%
\end{figure*}

\begin{figure*}[!htbp]
\centering
    \begin{minipage}[t]{0.45\linewidth}
        \centering
        \includegraphics[width=\textwidth]{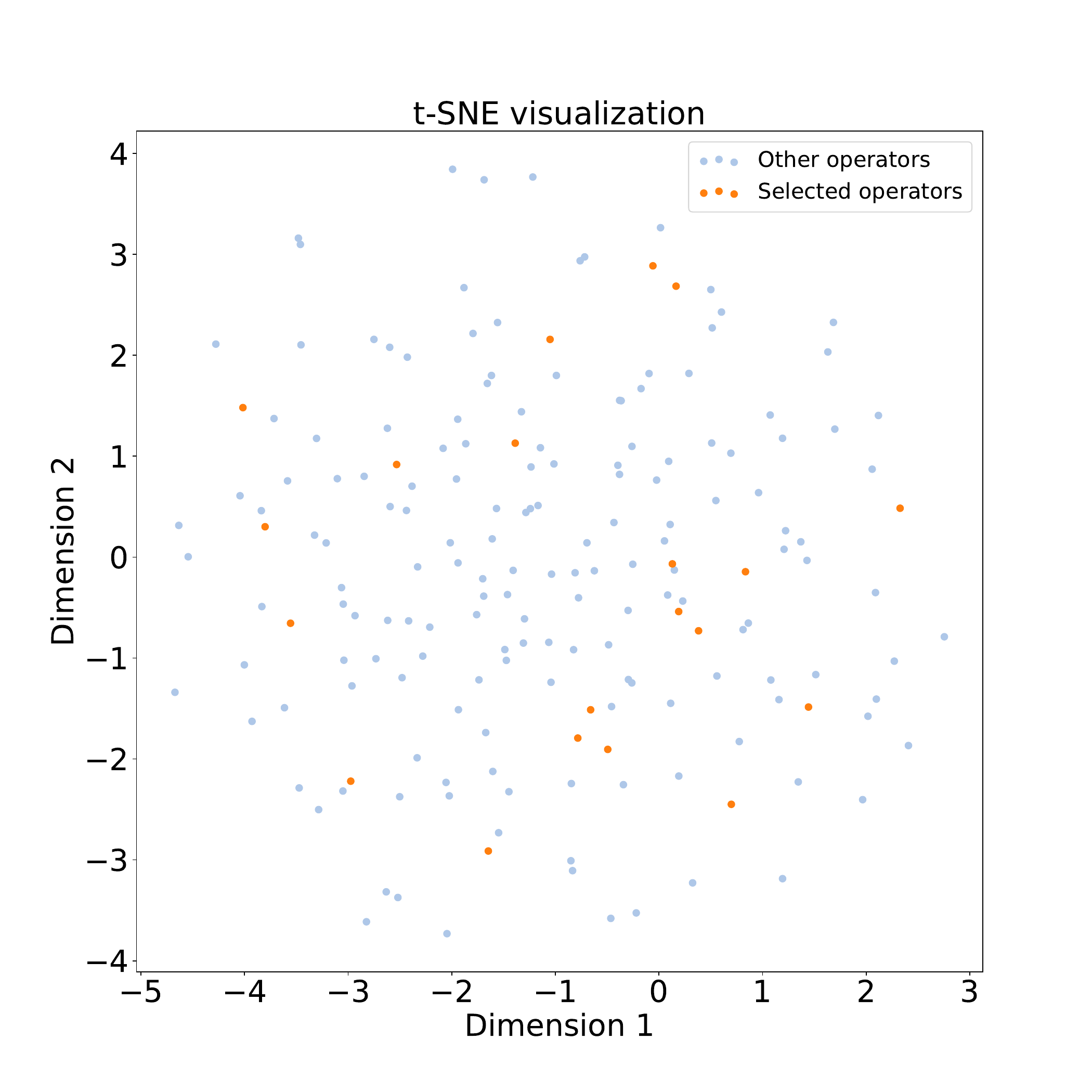}
        \vspace{-10pt}
        \captionsetup{font=scriptsize,justification=centering}\caption{\textbf{Node embeddings before training \\ (GAT - Operator w/o Attributes)}}
        \label{fig:gat_operator_before}
    \end{minipage}%
    \begin{minipage}[t]{0.45\linewidth}
        \centering
        \includegraphics[width=\textwidth]{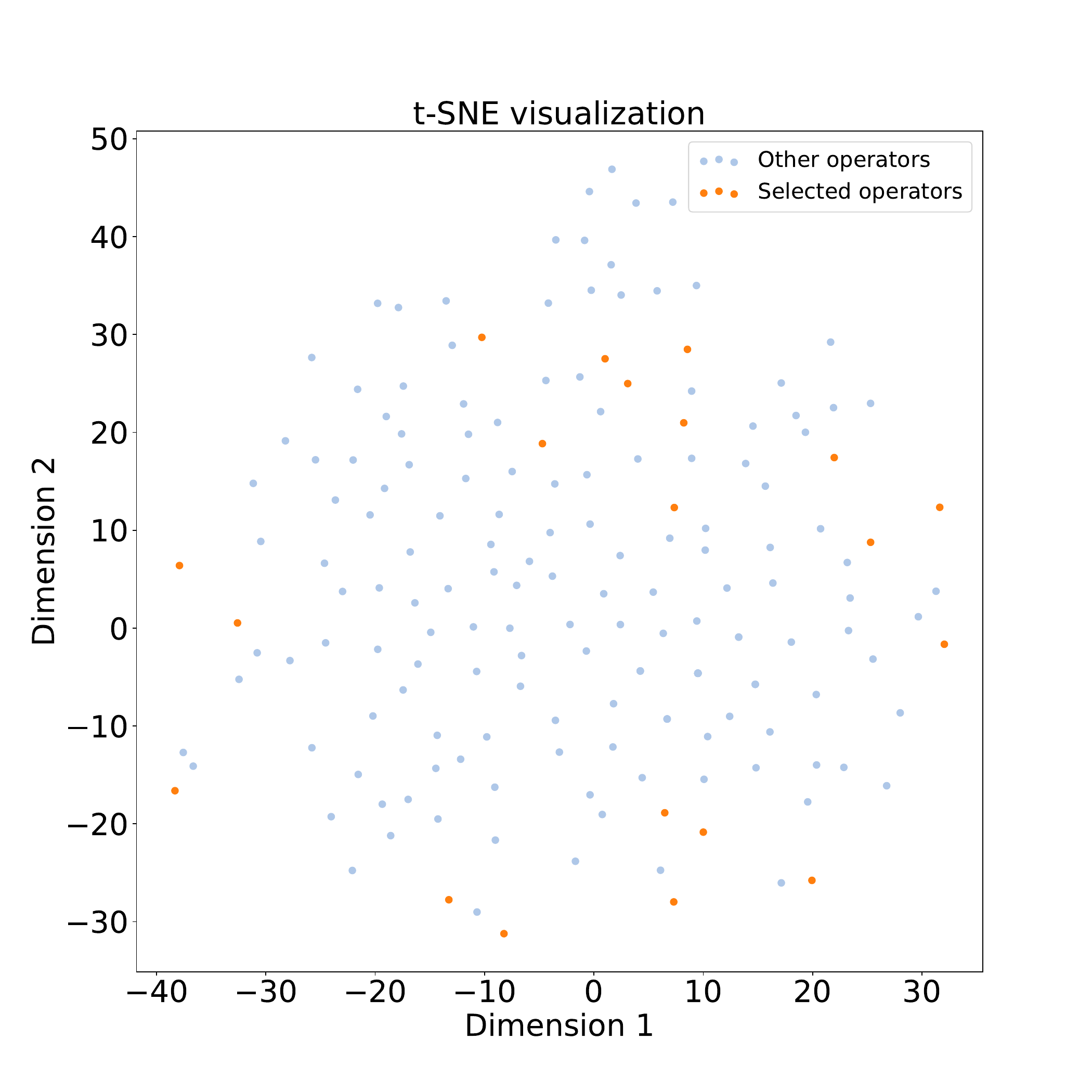}
        \vspace{-10pt}
        \captionsetup{font=scriptsize,justification=centering}\caption{\textbf{Node embeddings after training \\ (GAT - Operator w/o Attributes)}}
        \label{fig:gat_operator_after}
    \end{minipage}%
\end{figure*}

\begin{figure*}[!htbp]
\centering
    \begin{minipage}[t]{0.45\linewidth}
        \centering
        \includegraphics[width=\textwidth]{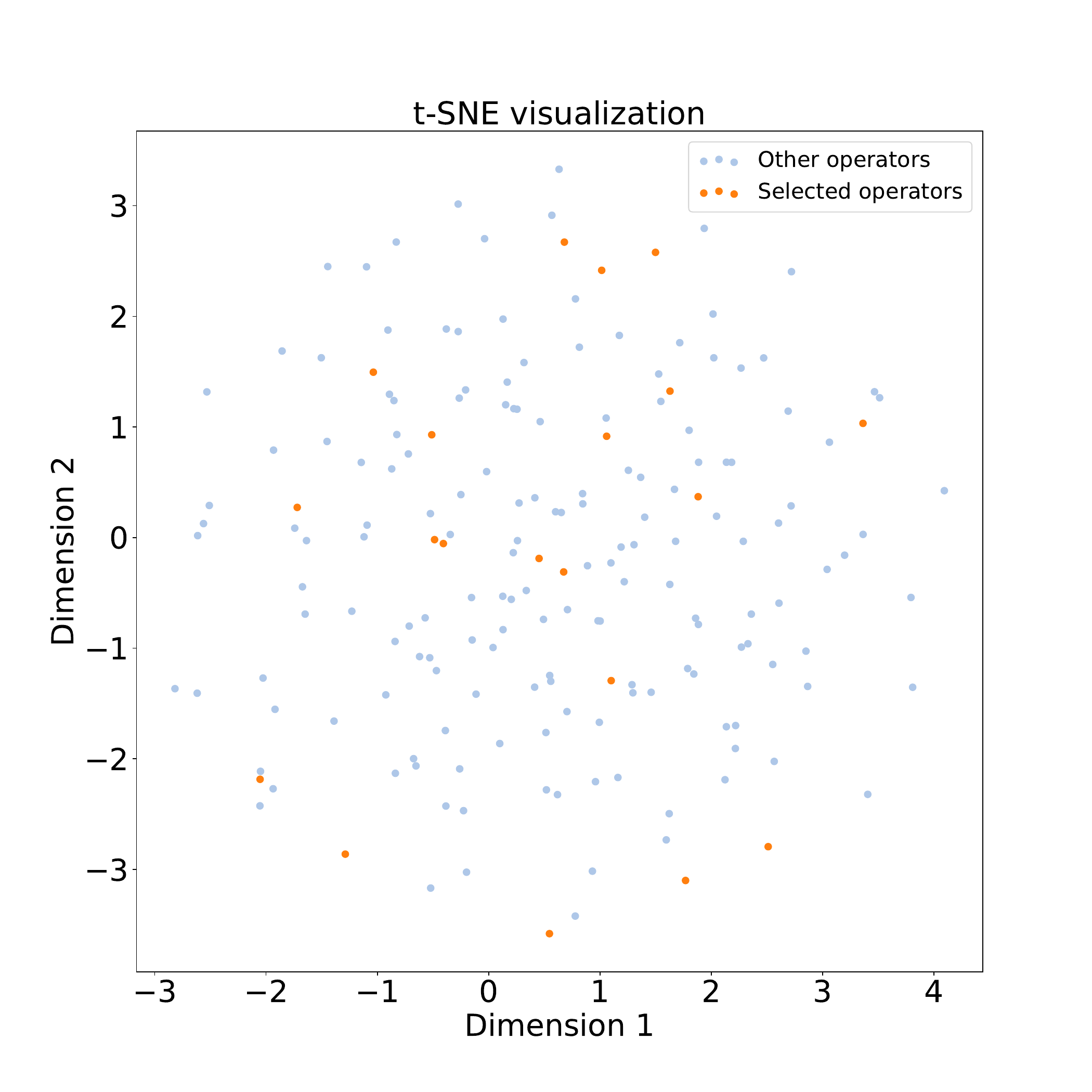}
        \vspace{-10pt}
        \captionsetup{font=scriptsize,justification=centering}\caption{\textbf{Node embeddings before training \\ (SAGE - Operator w/o Attributes)}}
        \label{fig:sage_operator_before}
    \end{minipage}%
    \begin{minipage}[t]{0.45\linewidth}
        \centering
        \includegraphics[width=\textwidth]{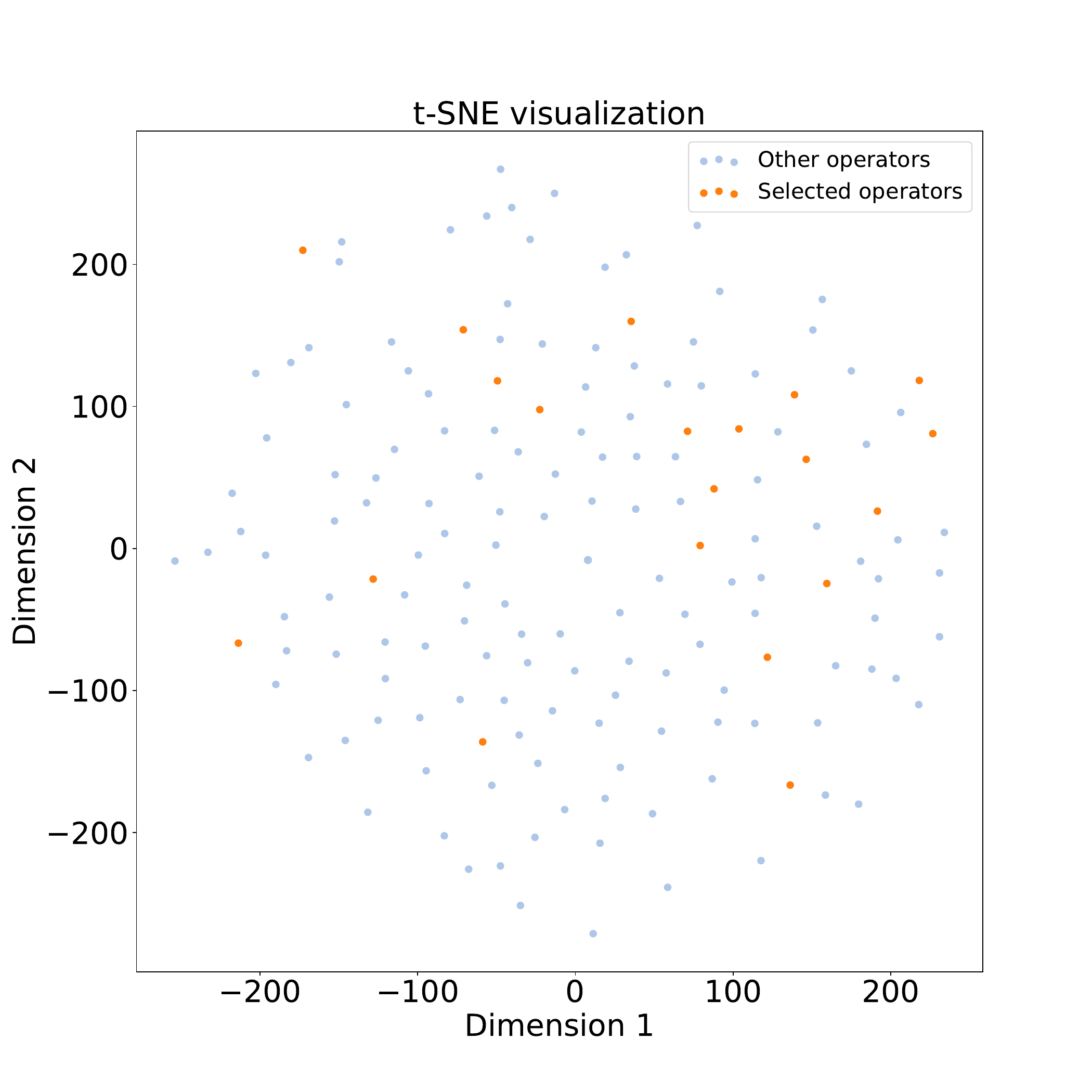}
        \vspace{-10pt}
        \captionsetup{font=scriptsize,justification=centering}\caption{\textbf{Node embeddings after training \\ (SAGE - Operator w/o Attributes)}}
        \label{fig:sage_operator_after}
    \end{minipage}%
\end{figure*}

Figure \ref{fig:gcn_operator_before}-\ref{fig:sage_operator_after} show the t-SNE visualization results of node embeddings before and after training from GCN, GAT, and SAGE with node features denoted as `Operator w/ Attributes.' The orange points represent Younger's top 20 most frequently occurring operators. It can be seen that the distribution of node embeddings is relatively concentrated before training, while the distribution of all embeddings is uniform after training. This result indicates baseline models learned the features of different nodes well.


\begin{figure*}[!htbp]
\centering
    \begin{minipage}[t]{0.45\linewidth}
        \centering
        \includegraphics[width=\textwidth]{figures/gcn_node_subgraph_embedding}
        \vspace{-10pt}
        \captionsetup{font=scriptsize,justification=centering}\caption{\textbf{Subgraph embeddings \\ (GCN - Operator w/ Attributes)}}
        \label{fig:gcn_node_subgraph}
    \end{minipage}%
    \begin{minipage}[t]{0.45\linewidth}
        \centering
        \includegraphics[width=\textwidth]{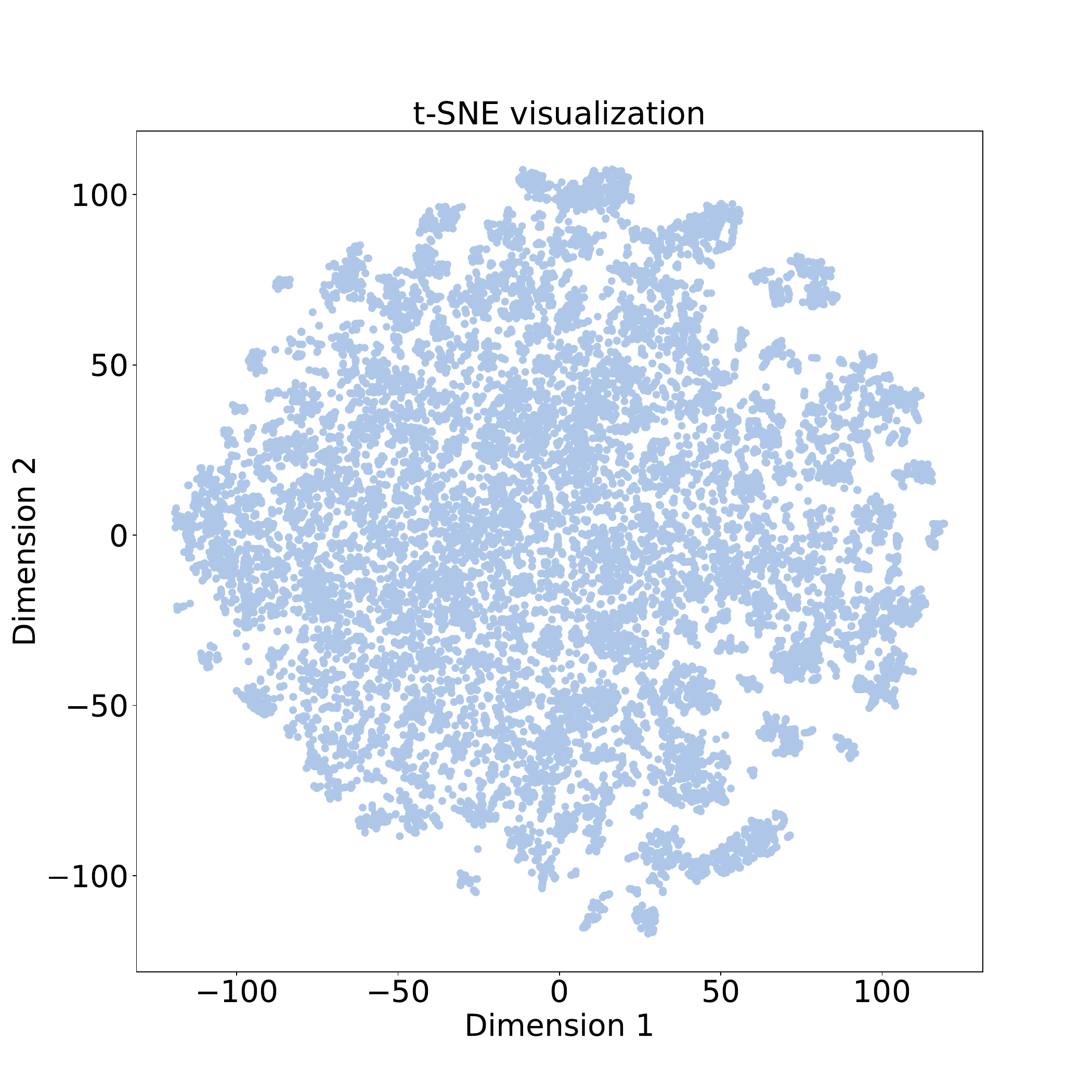}
        \vspace{-10pt}
        \captionsetup{font=scriptsize,justification=centering}\caption{\textbf{Subgraph embeddings \\ (GCN - Operator w/o Attributes)}}
        \label{fig:gcn_operator_subgraph}
    \end{minipage}%
\end{figure*}


\begin{figure*}[!htbp]
\centering
    \begin{minipage}[t]{0.45\linewidth}
        \centering
        \includegraphics[width=\textwidth]{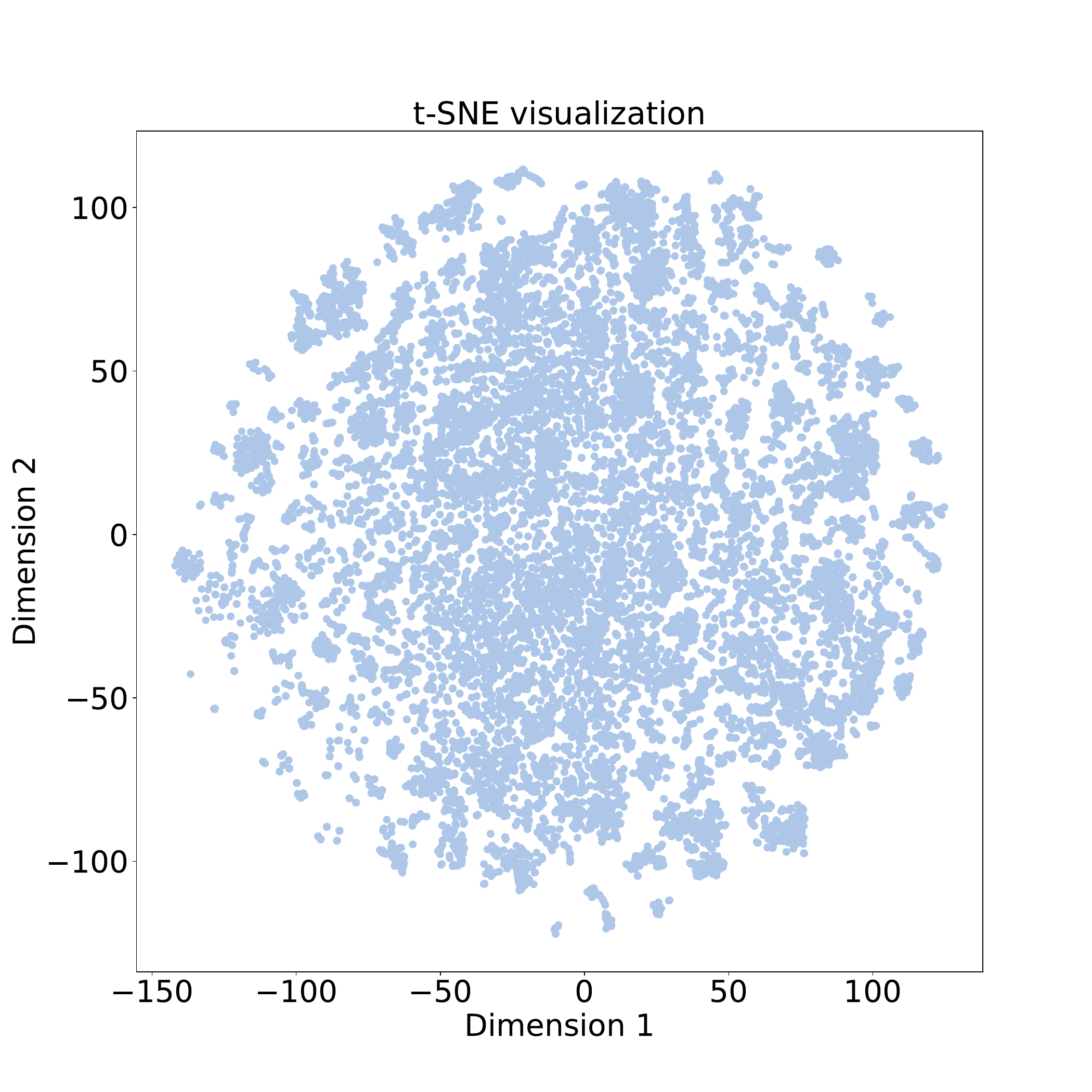}
        \vspace{-10pt}
        \captionsetup{font=scriptsize,justification=centering}\caption{\textbf{Subgraph embeddings \\ (SAGE - Operator w/ Attributes)}}
        \label{fig:sage_node_subgraph}
    \end{minipage}%
    \begin{minipage}[t]{0.45\linewidth}
        \centering
        \includegraphics[width=\textwidth]{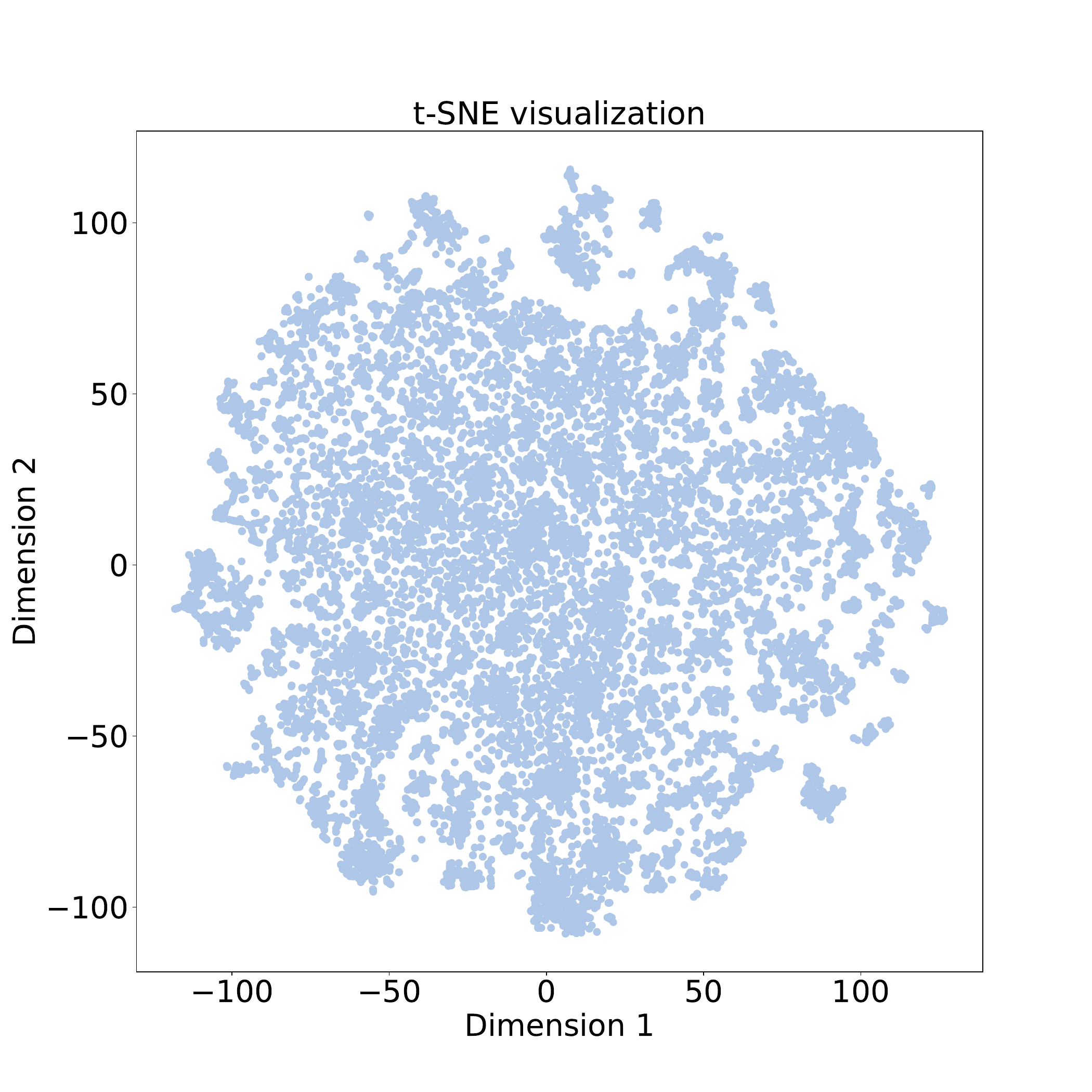}
        \vspace{-10pt}
        \captionsetup{font=scriptsize,justification=centering}\caption{\textbf{Subgraph embeddings \\ (SAGE - Operator w/o Attributes)}}
        \label{fig:sage_operator_subgraph}
    \end{minipage}%
\end{figure*}


\begin{figure*}[!htbp]
\centering
    \begin{minipage}[t]{0.45\linewidth}
        \centering
        \includegraphics[width=\textwidth]{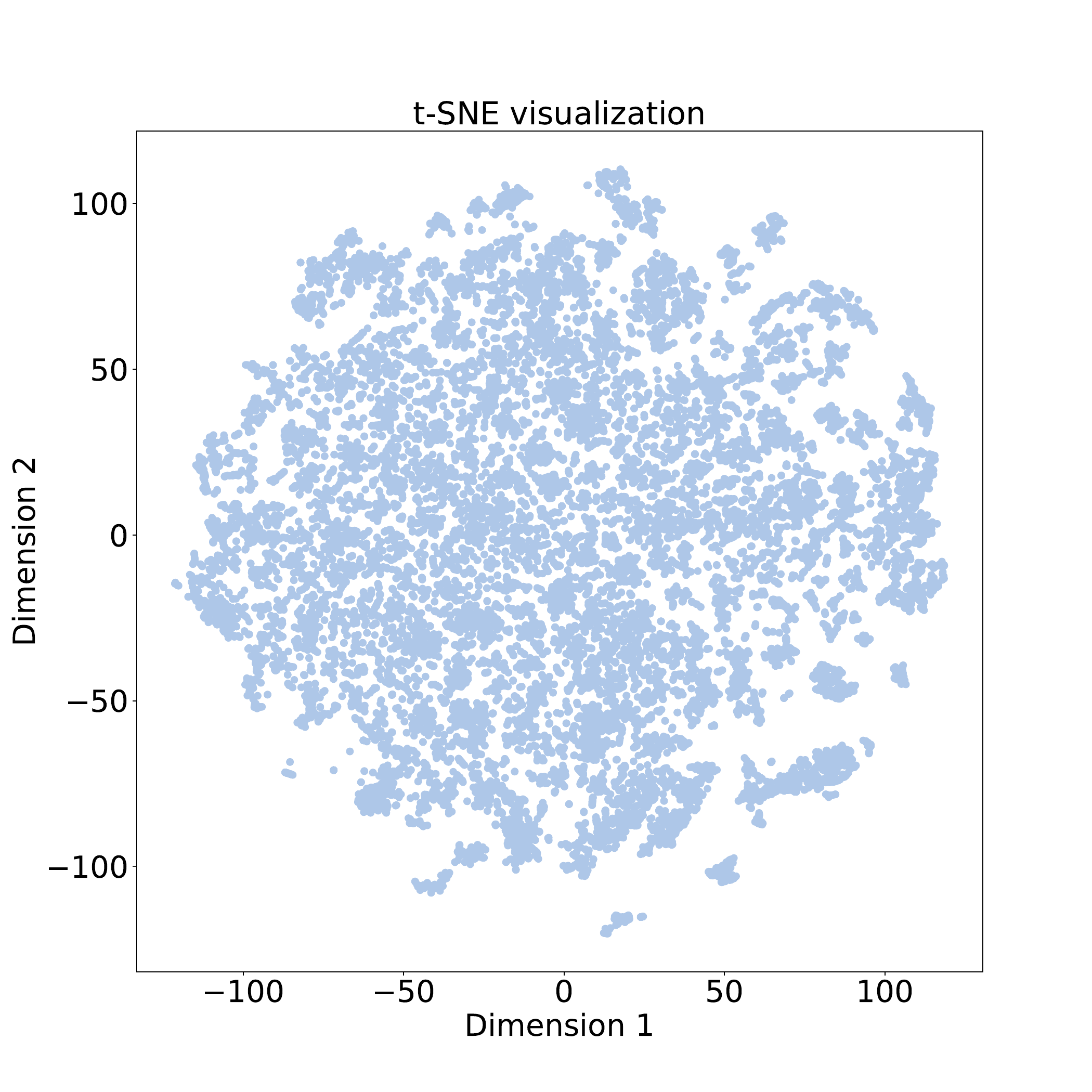}
        \vspace{-10pt}
        \captionsetup{font=scriptsize,justification=centering}\caption{\textbf{Subgraph embeddings \\ (GAT - Operator w/o Attributes)}}
        \label{fig:gat_operator_subgraph}
    \end{minipage}%
\end{figure*}



\begin{figure*}[!htbp]
\centering
    \begin{minipage}[t]{0.45\linewidth}
        \centering
        \includegraphics[width=\textwidth]{figures/gcn_node_graph_embedding}
        \vspace{-10pt}
        \captionsetup{font=scriptsize,justification=centering}\caption{\textbf{Graph embeddings \\ (GCN - Operator w/ Attributes)}}
        \label{fig:gcn_node_graph}
    \end{minipage}%
    \begin{minipage}[t]{0.45\linewidth}
        \centering
        \includegraphics[width=\textwidth]{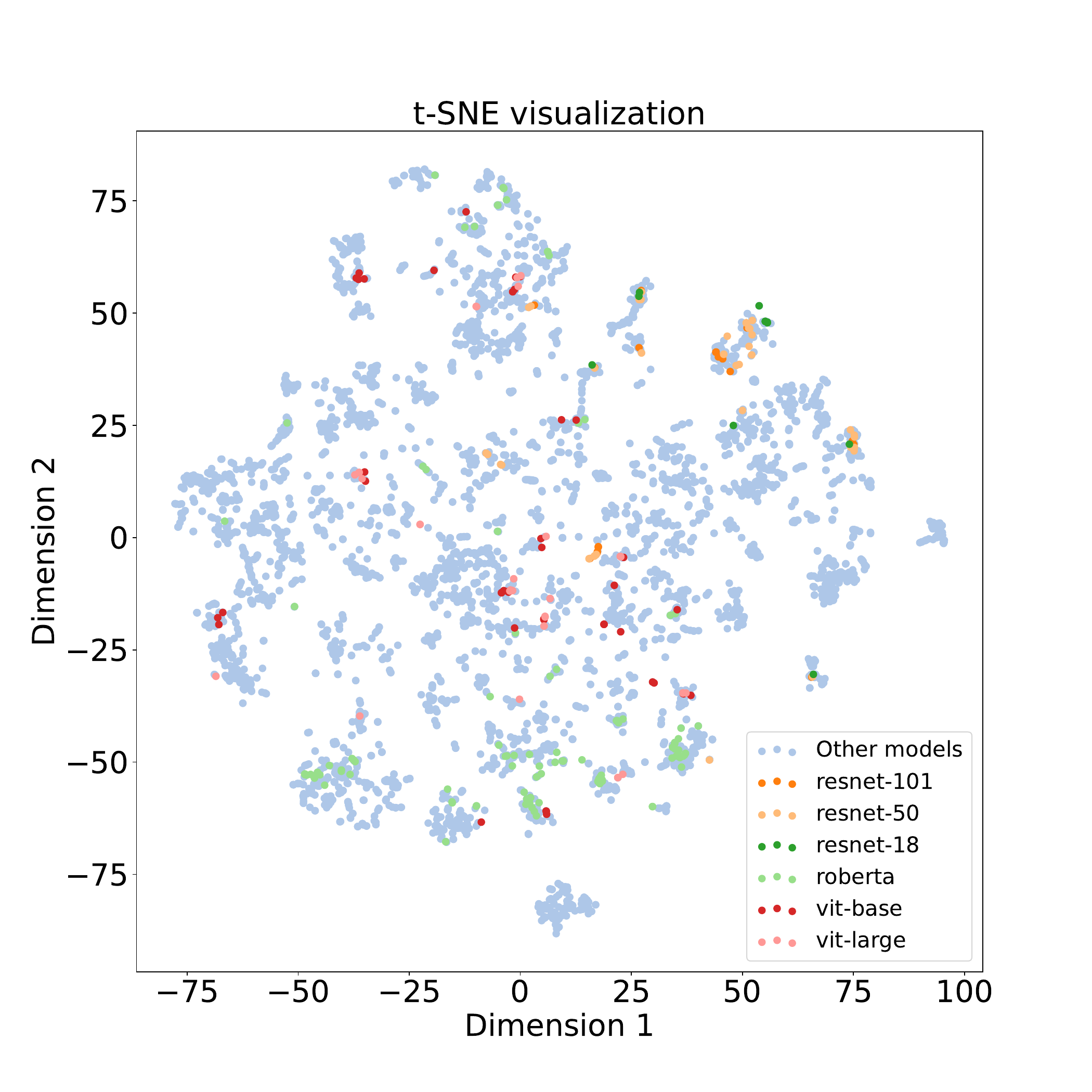}
        \vspace{-10pt}
        \captionsetup{font=scriptsize,justification=centering}\caption{\textbf{Graph embeddings \\ (GCN - Operator w/o Attributes)}}
        \label{fig:gcn_operator_graph}
    \end{minipage}%
\end{figure*}


\begin{figure*}[!htbp]
\centering
    \begin{minipage}[t]{0.45\linewidth}
        \centering
        \includegraphics[width=\textwidth]{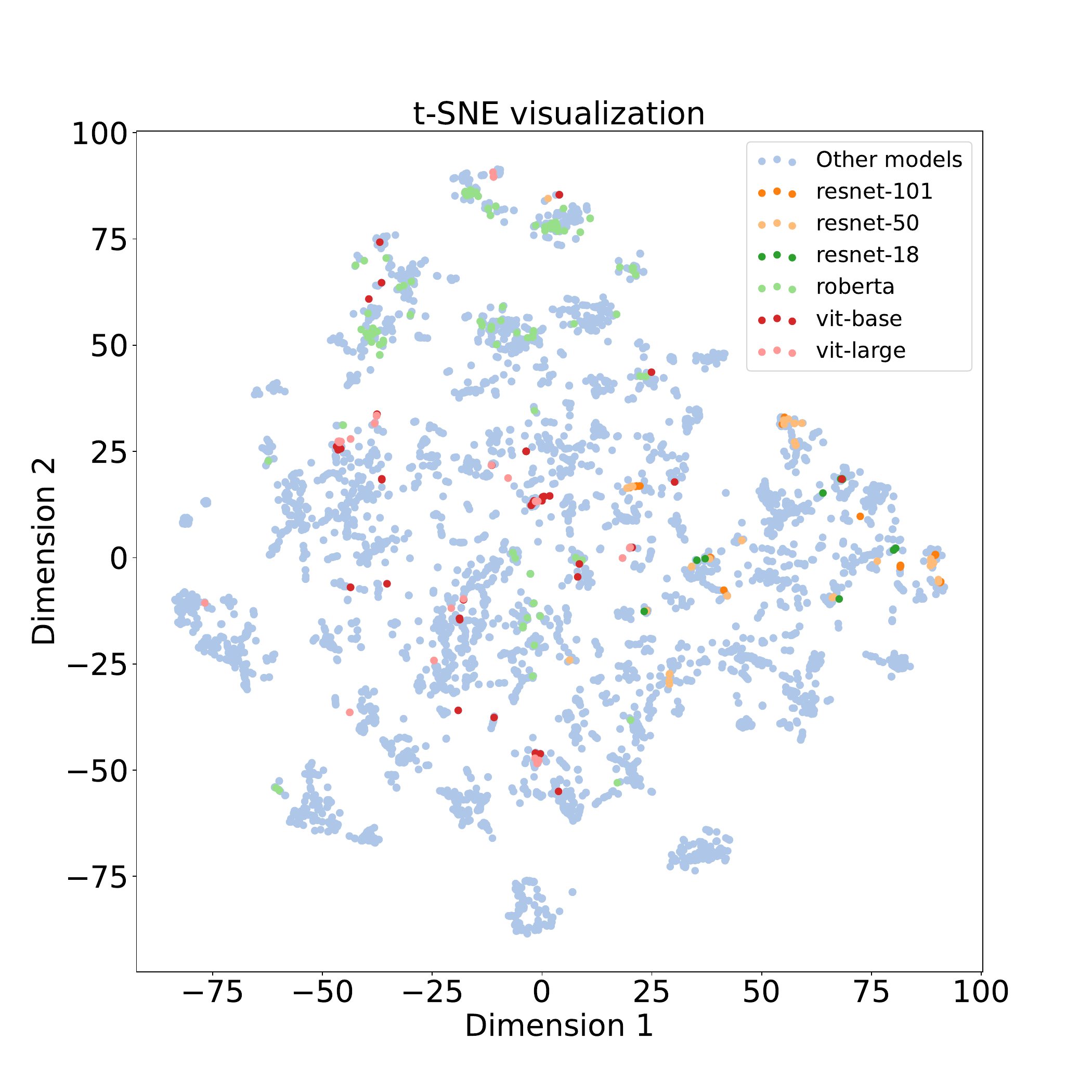}
        \vspace{-10pt}
        \captionsetup{font=scriptsize,justification=centering}\caption{\textbf{Graph embeddings \\ (SAGE - Operator w/ Attributes)}}
        \label{fig:sage_node_graph}
    \end{minipage}%
    \begin{minipage}[t]{0.45\linewidth}
        \centering
        \includegraphics[width=\textwidth]{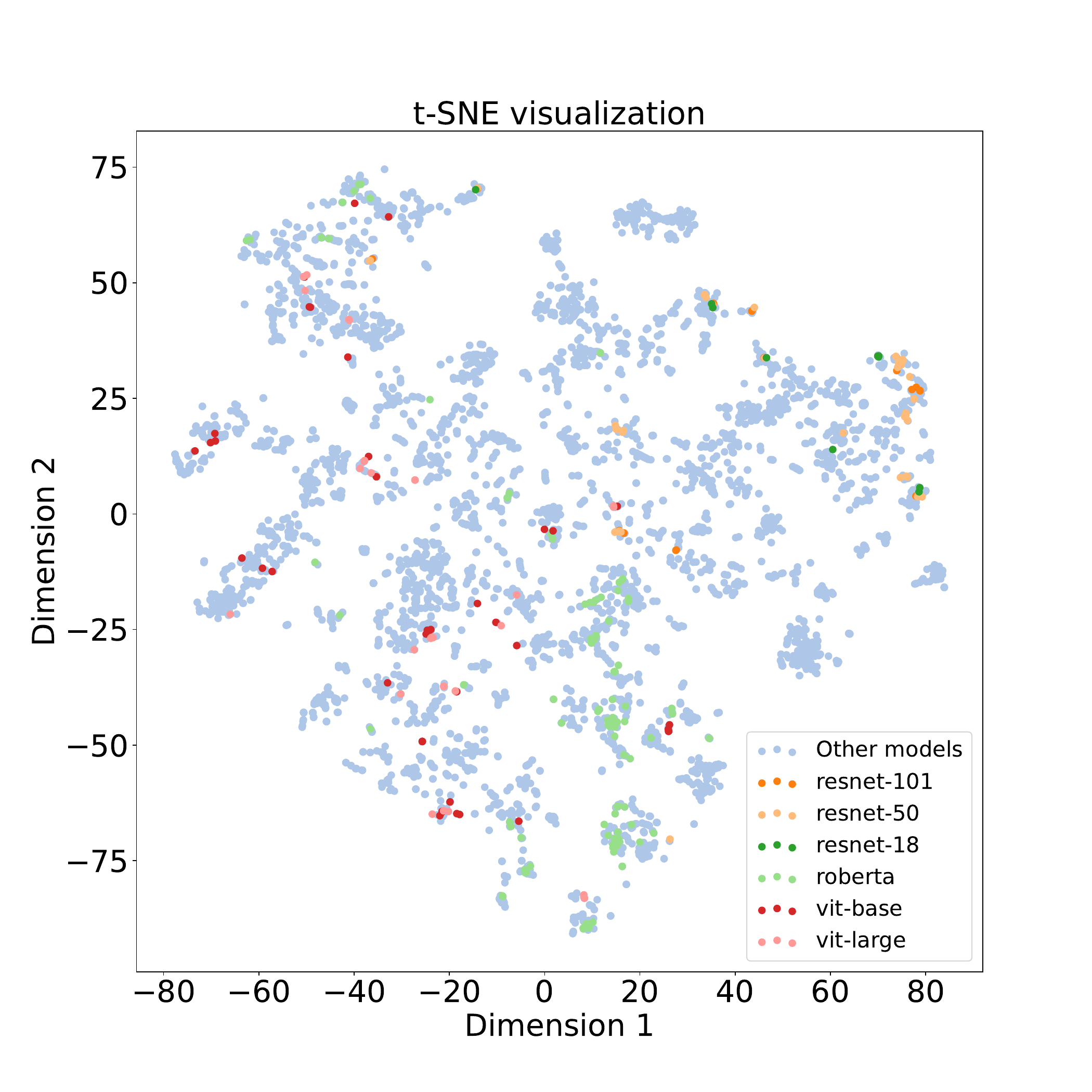}
        \vspace{-10pt}
        \captionsetup{font=scriptsize,justification=centering}\caption{\textbf{Graph embeddings \\ (SAGE - Operator w/o Attributes)}}
        \label{fig:sage_operator_graph}
    \end{minipage}%
\end{figure*}


\begin{figure*}[!htbp]
\centering
    \begin{minipage}[t]{0.45\linewidth}
        \centering
        \includegraphics[width=\textwidth]{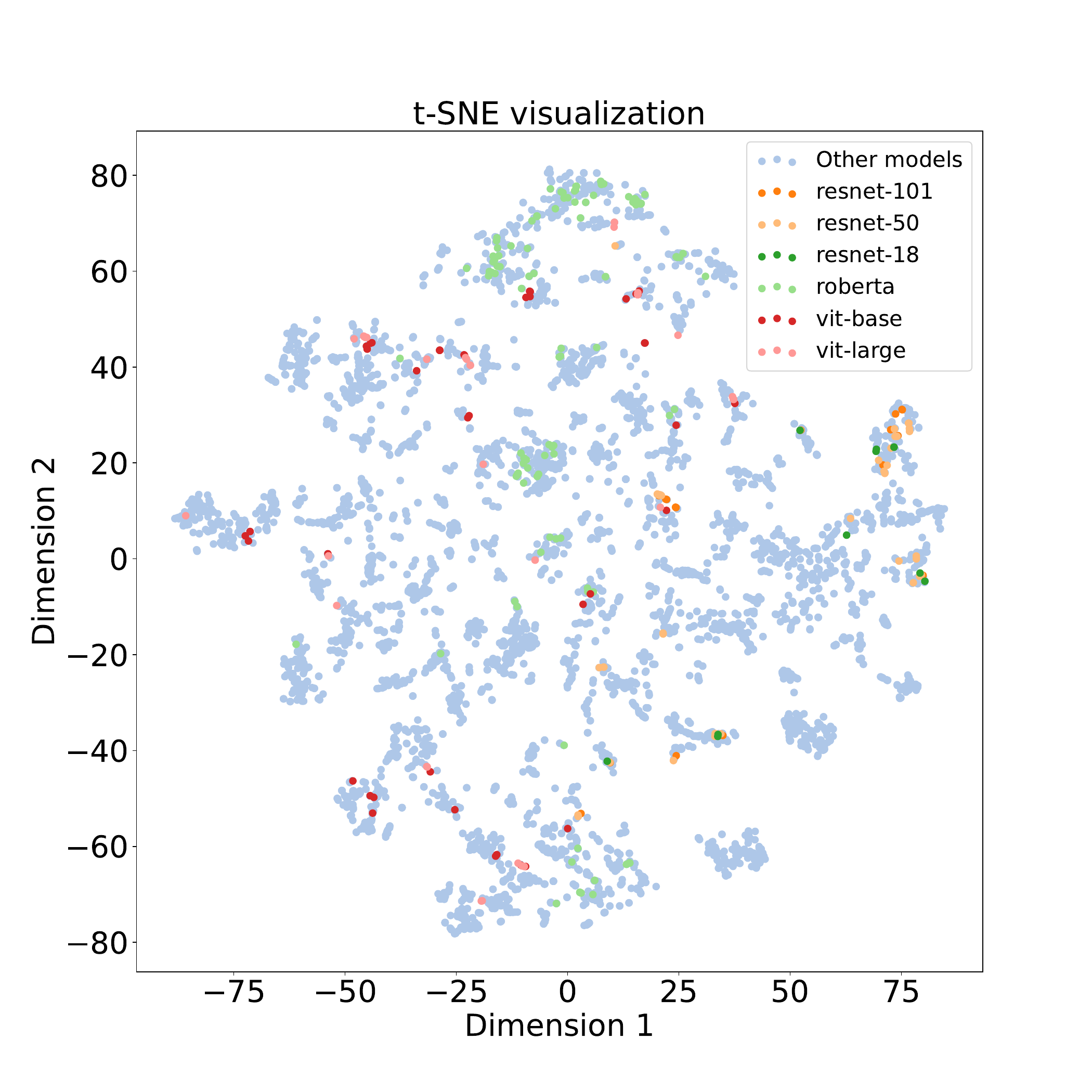}
        \vspace{-10pt}
        \captionsetup{font=scriptsize,justification=centering}\caption{\textbf{Graph embeddings \\ (SAGE - Operator w/o Attributes)}}
        \label{fig:gat_operator_graph}
    \end{minipage}%
\end{figure*}


\subsection{Subgraph Embedding}
\subsubsection{Checkpoint Selection}
To better illustrate the distribution of subgraphs in Younger in high-dimensional space, we selected checkpoints of baseline models according to the method from Section~\ref{apx:df_design_checkpoint}. Then, we calculated the embeddings of these subgraphs using operators with attributes and those without attributes.

\subsubsection{Visualization}
Figure \ref{fig:gcn_node_subgraph}-\ref{fig:gat_operator_subgraph} show the t-SNE visualization results of all subgraph embeddings under the GCN, GAT, and SAGE models. Due to memory overflow in `Operator w/ Attributes' of GAT, we only present the visualization of `Operator w/o Attributes' about GAT. As can be seen, all three models have distinguished the embeddings of subgraphs well, but due to data bias, node embeddings were not learned particularly well. Therefore, the models only distinguished the embeddings of subgraphs well in a part of the spatial distribution (the boundary of the space).
In addition, compared to `Operator w/o Attributes,' `Operator w/ Attributes' has a finer granularity in distinguishing subgraph embeddings, i.e., different clusters occupy less space.

\subsection{Graph Embedding}
\subsubsection{Obtaining Method}
We obtain the graph embeddings by averaging the embeddings of all subgraphs in each DAG.
Therefore, each baseline model can generate two types of graph embeddings: `Operator w/Attributes' and `Operator w/o Attributes.'
However, due to memory overflow in `Operator w/ Attributes' of GAT, we only present the visualization of `Operator w/o Attributes' about GAT.

\subsubsection{Visualization}
Figure \ref{fig:gcn_node_graph}-\ref{fig:gat_operator_graph} show the t-SNE visualization results of all graph embeddings under the GCN, GAT, and SAGE models. 
We mark the embeddings of several commonly used models in figures in different colors. Several architectures have shown almost similar results. It can be seen that, on the one hand, the embeddings of DAGs based on the same architecture are very close or even overlap in the graph; for example, there are many points of the RoBERTa~\cite{liu2019roberta} and ViT~\cite{dosovitskiy2020image} architectures, which are Transformer-based~\cite{vaswaniAttentionAllYou2017} architectures, that are close in distance or overlap. On the other hand, it can be seen that the Younger dataset covers multiple common architectures well, indicating that Younger covers most of the neural network architectures in the real world.
In addition, the same architecture has multiple points of the same color in the figures, indicating that the dataset contains various variants of this type of architecture.




\vspace{-10pt}
\section{Author Contributions}
\begin{itemize}
\item \textbf{Zhengxin Yang}: He led the project and proposed its idea. He independently implemented the project's code, including the dataset's construction code, the primary framework for application experiments, and statistical analysis.
\item \textbf{Jianfeng Zhan}: Proposed the concept of AIGNNA, participated in the entire project discussion, provided constructive suggestions, and provided guidance and supervision.
\item \textbf{Wanling Gao}: Participated in the discussion of the entire project, helped improve the idea of dataset application orientation, and provided constructive suggestions for application experiments.
\item \textbf{Luzhou Peng}: Participated in the debugging process of the code and jointly implemented a portion of the experimental code with Zhengxin Yang. He independently implemented some statistical analysis experimental code and partially participated in project discussions.
\item \textbf{Yunyou Huang}: He partially participated in project discussions and provided annotations on performance metric values, although Younger did not include this.
\item \textbf{Fei Tang}: Participated in discussions on the initial ideas of the project.
\end{itemize}
\end{appendices}

\newpage

\section*{Checklist}
\begin{enumerate}

\item For all authors...
\begin{enumerate}
  \item Do the main claims made in the abstract and introduction accurately reflect the paper's contributions and scope?
    \answerYes{}
  \item Did you describe the limitations of your work?
    \answerYes{In Section~\ref{experiments}}
  \item Did you discuss any potential negative societal impacts of your work?
    \answerNA{}
  \item Have you read the ethics review guidelines and ensured that your paper conforms to them?
    \answerYes{}
\end{enumerate}

\item If you are including theoretical results...
\begin{enumerate}
  \item Did you state the full set of assumptions of all theoretical results?
    \answerNA{}
	\item Did you include complete proofs of all theoretical results?
    \answerNA{}
\end{enumerate}

\item If you ran experiments (e.g. for benchmarks)...
\begin{enumerate}
  \item Did you include the code, data, and instructions needed to reproduce the main experimental results (either in the supplemental material or as a URL)?
    \answerYes{See Appendix~\ref{apx:exp}}
  \item Did you specify all the training details (e.g., data splits, hyperparameters, how they were chosen)?
    \answerYes{See Appendix~\ref{apx:exp}}
	\item Did you report error bars (e.g., with respect to the random seed after running experiments multiple times)?
    \answerNA{}
	\item Did you include the total amount of compute and the type of resources used (e.g., type of GPUs, internal cluster, or cloud provider)?
    \answerYes{See Appendix~\ref{apx:exp}}
\end{enumerate}

\item If you are using existing assets (e.g., code, data, models) or curating/releasing new assets...
\begin{enumerate}
  \item If your work uses existing assets, did you cite the creators?
    \answerYes{We provide citations for all resources and baselines}
  \item Did you mention the license of the assets?
    \answerYes{See Appendix~\ref{apx:lcs}}
  \item Did you include any new assets either in the supplemental material or as a URL?
    \answerNA{}
  \item Did you discuss whether and how consent was obtained from people whose data you're using/curating?
    \answerNA{}
  \item Did you discuss whether the data you are using/curating contains personally identifiable information or offensive content?
    \answerNA{}
\end{enumerate}

\item If you used crowdsourcing or conducted research with human subjects...
\begin{enumerate}
  \item Did you include the full text of instructions given to participants and screenshots, if applicable?
    \answerNA{}
  \item Did you describe any potential participant risks, with links to Institutional Review Board (IRB) approvals, if applicable?
    \answerNA{}
  \item Did you include the estimated hourly wage paid to participants and the total amount spent on participant compensation?
    \answerNA{}
\end{enumerate}

\end{enumerate}


\end{document}